\documentclass{article}

\usepackage[preprint]{neurips_2025}

\usepackage[normalem]{ulem}
\usepackage{multirow}
\usepackage{graphicx}
\usepackage{booktabs}
\usepackage[most]{tcolorbox}
\usepackage{fvextra} 

\usepackage{booktabs}
\usepackage{multirow}

\usepackage[utf8]{inputenc}
\usepackage{booktabs}
\usepackage{multirow}
\usepackage{hyperref}
\usepackage{xcolor}
\usepackage{geometry}
\usepackage{amsmath}
\usepackage{graphicx}
\usepackage[most]{tcolorbox}
\definecolor{rawbg}{HTML}{F4F6F8}
\definecolor{rawframe}{HTML}{78909C}

\definecolor{skillbg}{HTML}{EDF7ED}
\definecolor{skillframe}{HTML}{5C8A5C}

\definecolor{draftbg}{HTML}{EEF4FB}
\definecolor{draftframe}{HTML}{5E81AC}

\definecolor{promptbg}{HTML}{F7F7F7}
\definecolor{promptframe}{HTML}{757575}

\newtcolorbox{rawbox}[1][]{
    colback=rawbg, colframe=rawframe,
    fonttitle=\bfseries\small, title={#1},
    boxrule=0.5pt, arc=2pt,
    left=6pt, right=6pt, top=4pt, bottom=4pt,
    fontupper=\small
}

\newtcolorbox{skillbox}[1][]{
    colback=skillbg, colframe=skillframe,
    fonttitle=\bfseries\small, title={#1},
    boxrule=0.5pt, arc=2pt,
    left=6pt, right=6pt, top=4pt, bottom=4pt,
    fontupper=\small
}

\newtcolorbox{draftbox}[1][]{
    colback=draftbg, colframe=draftframe,
    fonttitle=\bfseries\small, title={#1},
    boxrule=0.5pt, arc=2pt,
    left=6pt, right=6pt, top=4pt, bottom=4pt,
    fontupper=\small
}

\newtcolorbox{promptbox}[1][]{
    colback=promptbg, colframe=promptframe,
    fonttitle=\bfseries\small, title={#1},
    boxrule=0.5pt, arc=2pt,
    left=6pt, right=6pt, top=4pt, bottom=4pt,
    fontupper=\small
}

\usepackage{booktabs}
\usepackage{multirow}
\usepackage{hyperref}
\usepackage{xcolor}
\usepackage{geometry}
\usepackage{amsmath}
\usepackage{graphicx}
\usepackage[most]{tcolorbox}
\usepackage{amsmath}
\usepackage{enumitem}
\usepackage[utf8]{inputenc} % allow utf-8 input
\usepackage[T1]{fontenc}    % use 8-bit T1 fonts
\usepackage{hyperref}       % hyperlinks
\usepackage{url}            % simple URL typesetting
\usepackage{booktabs}       % professional-quality tables
\usepackage{amsfonts}       % blackboard math symbols
\usepackage{nicefrac}       % compact symbols for 1/2, etc.
\usepackage{microtype}      % microtypography
\usepackage{xcolor}         % colors

\usepackage{hyperref}
\usepackage{fontawesome5}
\usepackage{graphicx}

\title{Skill Retrieval Augmentation for Agentic AI}

\usepackage{authblk}
\author[1, 2]{\textbf{Weihang Su}\textsuperscript{*}}
\author[1]{\textbf{Jianming Long}}
\author[1]{\textbf{Qingyao Ai}\textsuperscript{$\dagger$}}
\author[2]{\textbf{Qiaozhi He}}
\author[1]{\textbf{Yichen Tang}}
\author[1]{\textbf{Changyue Wang}}
\author[1]{\textbf{Yiteng Tu}}
\author[2]{\textbf{Yingbo Wang}}
\author[1]{\textbf{Yiqun Liu}}

\affil[1]{Department of Computer Science and Technology, Tsinghua University}
\affil[2]{ByteDance Inc.}

\begin{document}

\maketitle

\begingroup
\renewcommand{\thefootnote}{\fnsymbol{footnote}}
\footnotetext[1]{Work done during an internship at ByteDance. Email: oneal2000@126.com }
\footnotetext[2]{Corresponding author: aiqy@tsinghua.edu.cn}
\endgroup

\hypersetup{
  hidelinks
}

\vspace{-10mm}

% \begin{center}
% \small
% \href{https://github.com/oneal2000/SR-Agents}{\faIcon{github} GitHub: \url{https://github.com/oneal2000/SR-Agents}}
% \quad
% \\
% \vspace{2mm}
% \href{https://huggingface.co/datasets/WeihangSu/SRA-Bench}{
%   \raisebox{-0.15em}{\includegraphics[height=1em]{pic/huggingface_original.png}}
%   Hugging Face: \url{https://huggingface.co/datasets/WeihangSu/SRA-Bench}
% }
% \end{center}

\begin{center}
\small

\href{https://github.com/oneal2000/SR-Agents}{\faIcon{github} GitHub: \url{https://github.com/oneal2000/SR-Agents}}
\\
\vspace{1mm}

\href{https://sr-agents.github.io}{\faIcon{link} Project Website: \url{https://sr-agents.github.io}}
\\
\vspace{1mm}

\href{https://huggingface.co/datasets/WeihangSu/SRA-Bench}{
  \raisebox{-0.15em}{\includegraphics[height=1em]{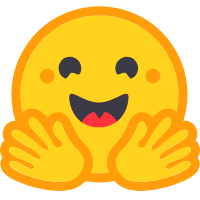}}
  Hugging Face: \url{https://huggingface.co/datasets/WeihangSu/SRA-Bench}
}
\end{center}

\vspace{1mm}

\begin{abstract}

As large language models (LLMs) evolve into agentic problem solvers, they increasingly rely on external, reusable skills to handle tasks beyond their native parametric capabilities. In existing agent systems, the dominant strategy for incorporating skills is to explicitly enumerate available skills within the context window. However, this strategy fails to scale: as skill corpora expand, context budgets are consumed rapidly, and the agent becomes markedly less accurate in identifying the right skill. To this end, this paper formulates Skill Retrieval Augmentation (SRA), a new paradigm in which agents dynamically retrieve, incorporate, and apply relevant skills from large external skill corpora on demand. To make this problem measurable, we construct a large-scale skill corpus and introduce \textbf{SRA-Bench}, the first benchmark for decomposed evaluation of the full SRA pipeline, covering skill retrieval, skill incorporation, and end-task execution. SRA-Bench contains 5,400 capability-intensive test instances and 636 manually constructed gold skills, which are mixed with web-collected distractor skills to form a large-scale corpus of 26,262 skills. Extensive experiments show that retrieval-based skill augmentation can substantially improve agent performance, validating the promise of the paradigm. At the same time, we uncover a fundamental gap in skill incorporation: current LLM agents tend to load skills at similar rates, regardless of whether a gold skill is retrieved or whether the task actually requires external capabilities. This shows that the bottleneck in skill augmentation lies not only in retrieval but also in the base model’s ability to determine which skill to load and when external loading is actually needed. These findings position SRA as a distinct research problem and establish a foundation for the scalable augmentation of capabilities in future agent systems. {All the code and data are released at the following GitHub link: \url{https://github.com/oneal2000/SR-Agents}}.

\end{abstract}

\section{Introduction}

Recent advancements in Large Language Models (LLMs) have catalyzed a paradigm shift toward Agentic AI, where models are evolving from passive text generators into active problem solvers capable of reasoning, planning, tool calling, and interacting with external environments~\cite{wang2024surveyagent,yao2022react,schick2023toolformer,shen2023hugginggpt}.
As these agents are expected to solve broader and more complex tasks, their internal parametric knowledge is no longer sufficient to support robust performance~\cite{wang2023voyager,karpas2022mrkl}. 
Instead, tackling such diverse and open-ended tasks increasingly relies on externalized, reusable capabilities that extend an agent's competence beyond the base model's intrinsic capacity.
In many emerging systems (e.g., OpenClaw~\cite{steinberger2026openclaw}), these capabilities are packaged as \textbf{\emph{skills}}: modular capability packages that help an agent solve recurring classes of problems, often encompassing natural-language instructions, invocation conditions, tool-usage procedures, executable code, and auxiliary resources~\cite{li2026skillsbench,ling2026agentskills}.

However, the prevailing paradigm for equipping agents with such capabilities relies on a fundamentally unscalable mechanism: explicit in-context skill injection~\cite{steinberger2026openclaw,openai2026codex,anthropic2026claudecode}. 
Current frameworks typically construct prompts that enumerate available skills, often compressing them into compact metadata or instructional summaries (e.g., the SKILL.md files used in OpenClaw~\cite{steinberger2026openclaw}), and then require the LLM to evaluate numerous candidates in context to identify and load relevant skills on demand based on user instructions.
While this design is intuitive and aligns naturally with existing agentic workflows, it becomes increasingly impractical in real-world deployments as the scale of skill libraries continues to grow~\cite{skillsmp2026}. 
In the past few months, the ecosystem of available skills has been expanding explosively: as of April 26, 2026, platforms such as SkillsMP~\cite{skillsmp2026} host more than one million distinct skills, and agents are expected to maintain large and lifelong skill libraries that continue to grow~\cite{ai2025memorybench,wang2026improve}.
Under this setting, the conventional practice of exposing all candidate skills in context begins to collapse under two compounding bottlenecks: the hard architectural limits of context windows, and the sharp degradation in reasoning and selection accuracy when the model is confronted with a massive volume of skills. 
Consequently, dynamically retrieving and injecting the right capability from a vast, out-of-context skill collection has emerged as a critical research problem.

To address this scalability failure, we formulate \textbf{Skill Retrieval Augmentation (SRA)}, a new paradigm for augmenting agents with external capabilities at scale. 
Rather than relying on a small, fixed set of visible skills in the prompt, SRA treats skills as entries in a large external capability corpus and requires agents to retrieve, load, and apply relevant skills on demand. 
Under this paradigm, we propose \textbf{Skill Retrieval Augmented Agents (SR-Agents)}, which dynamically retrieve and use relevant skills from large-scale skill corpora to expand their problem-solving capabilities. SRA is closely related to Retrieval-Augmented Generation (RAG), but it is not simply RAG with a different retrieval target. In classical knowledge-centric RAG, retrieved items are primarily declarative evidence used to ground generation. 
In contrast, SRA retrieves executable capabilities that augment the agent’s functional competence, rather than merely providing declarative knowledge to support generation.
Accordingly, retrieval in SRA should be evaluated not only by semantic relevance, but also by downstream utility: whether the retrieved candidate set contains the skills relevant to the current task, whether those skills can be correctly incorporated into the agent’s problem-solving process, and whether their application ultimately improves end-task performance.
This difference fundamentally changes the problem structure. Beyond skill retrieval itself, SRA introduces a multi-stage pipeline with three tightly coupled components: {skill retrieval}, which asks whether the system can identify relevant skills for a given user request from a large external corpus; {skill incorporation}, which asks whether the agent can correctly recognize, organize, and invoke the truly useful skills among retrieved candidates without being distracted or overwhelmed by irrelevant ones; and {skill application}, which asks whether incorporated skills are actually leveraged during task solving in ways that improve agent behavior and end-task success. Together, these components define a broader research agenda for scalable capability augmentation in agent systems. More broadly, this paradigm opens up new directions in skill indexing and organization, quality control over heterogeneous skills, and feedback-driven skill debugging, refinement, and lifelong accumulation.

To make this paradigm scientifically tractable, we build a large-scale {skill corpus} and introduce \textbf{SRA-Bench}, the first benchmark specifically designed for studying the SRA paradigm. 
Each instance in SRA-Bench is annotated with a user query, a ground-truth answer, and the corresponding gold skill(s), enabling fine-grained, decomposed evaluation across the full SRA pipeline. 
Rather than asking whether an agent ultimately succeeds, this benchmark allows us to diagnose three distinct questions: whether the system can retrieve the right skills from a large external corpus, whether the agent can correctly identify and incorporate the truly useful skills among retrieved candidates, and whether such retrieved skills can be translated into improved behavior and end-task performance. 
In this sense, the contribution of SRA-Bench is not merely to introduce a new benchmark, but to formulate skill retrieval augmentation as a concrete research problem that can be clearly defined, rigorously evaluated, and systematically analyzed.

Through extensive experiments on SRA-Bench, we uncover several key empirical findings about scalable skill augmentation.
First, even a simple SRA pipeline that retrieves a single skill using a general-purpose retriever and injects it into the context can already improve strong LLM agents over their skill-free counterparts, establishing the practical promise of our proposed paradigm.
Beyond this positive result, however, our experiments reveal that scalable skill augmentation is bottlenecked not only by retrieval quality, but by a previously underappreciated variable: whether the agent chooses to load external skills at all.
We find that skill-loading behavior is highly model-dependent: using the same skill corpus and retriever, different models exhibit dramatically different loading rates, with no monotonic trend that larger models behave more rationally or robustly.
Moreover, agents attempt to load skills at nearly identical rates regardless of whether the gold skill is actually present among the retrieved candidates, revealing a profound disconnect between successful retrieval and effective utilization.
More importantly, agents are no more likely to invoke a skill on tasks that genuinely require external capability than on tasks they can already solve natively, suggesting a fundamental absence of need-aware skill invocation.
Taken together, these findings show that scalable skill augmentation is not merely a retrieval problem, but a broader challenge of controlled skill exposure, need-aware incorporation, and reliable application, thereby motivating SRA as a distinct research agenda for agent systems.

To summarize, our contributions are as follows:
\begin{itemize}[leftmargin=*]

\item We propose {Skill Retrieval Augmentation (SRA)}, a new paradigm in which LLM-based agents retrieve and operationalize reusable external skills from large-scale skill corpora rather than relying on small, fixed skill sets directly placed in context.
    
    \item We build a large-scale {skill corpus} and introduce {SRA-Bench}, a benchmark resource with query, answer, and gold-skill annotations that supports decomposed evaluation of skill retrieval, skill incorporation, and end-to-end task solving.
    
    \item We establish {SR-Agents} as a baseline family for studying the SRA pipeline and conduct a systematic empirical analysis across models, retrievers, and task domains.
    
    \item We provide empirical findings that clarify both the promise and the bottlenecks of scalable skill use, showing that while external skills can substantially improve agent performance, effective access to capabilities requires advances not only in retrieval but also in skill incorporation and application.
    
\end{itemize}

\begin{figure*}[t]
\centering
    \includegraphics[width=\textwidth]{pic/Overall.pdf}
    \caption{An illustration of the Skill Retrieval Augmentation (SRA) paradigm. The agent retrieves candidate skills from a large external skill corpus, selectively incorporates useful skills into context, and applies them for downstream reasoning and acting. Black arrows denote the standard SRA workflow, while blue arrows represent iterative skill retrieval during reasoning and acting.} 
    \label{pic_overall}
\end{figure*}

\section{Problem Formulation}

Building upon the conceptual foundation above, we now formalize the core objects and operational pipeline of {Skill Retrieval Augmentation (SRA)}. 
Our formulation characterizes scalable skill augmentation as a multi-stage process that separates whether an agent can retrieve relevant skills, correctly incorporate them into its active problem-solving state, and ultimately translate them into improved task performance.

\subsection{Agent Skills and Skill Corpus}

We begin by defining the basic unit of SRA: the {agent skill}. 
Unlike ordinary retrieved documents, which primarily provide declarative evidence for grounded generation, or standalone tool APIs, which expose only isolated callable interfaces, a skill is a reusable capability package that enables an agent to solve a recurring class of problems. 
A standard skill typically contains a name, a short natural-language description, detailed usage instructions, invocation conditions, procedural guidance, executable code, and auxiliary resources, thereby exposing not only \emph{what} capability is available, but also \emph{when} and \emph{how} it should be applied. 

Formally, let $\mathcal{C} = \{s_1, s_2, \dots, s_N\}$ denote a {skill corpus} containing $N$ skills. 
Each skill $s_i \in \mathcal{C}$ is represented as

\begin{equation}
    s_i = (n_i, r_i, c_i, \pi_i),
\end{equation}

where $n_i$ is the \textbf{name} of the skill, serving as its compact semantic identifier, $r_i$ is a short \textbf{description} summarizing the capability and intended use of the skill, and $c_i$ is the main \textbf{content}, which may include natural-language instructions, usage constraints, procedural guidance, and other language-visible artifacts exposed to the agent. 
The executable payload $\pi_i$ includes code, tools, or other operational resources that realize the capability in an external environment.

In practical systems, these components may correspond to artifacts such as a skill title, a one-line summary, \texttt{SKILL.md}, scripts, and static assets. 
More generally, a skill is any modular unit that couples a natural-language interface with an executable capability.

\subsection{Skill Retrieval Augmentation}

Given a user query $q$ and a large external skill corpus $\mathcal{C}$, the objective of SRA is to augment an agent with relevant external capabilities retrieved from a large skill corpus on demand, rather than relying on a pre-exposed, fixed set of skills enumerated directly in the context.
We formalize this process as a three-stage pipeline.

\paragraph{Skill Retrieval.}
A retriever $R$ first maps the user query $q$ and the skill corpus $\mathcal{C}$ to a ranked list of candidate skills:
\begin{equation}
\mathcal{L}_k = R(q, \mathcal{C}) = [s^{(1)}, s^{(2)}, \dots, s^{(k)}], \qquad s^{(j)} \in \mathcal{C}, \; k \ll N.
\end{equation}
The retrieved candidates are ranked by their estimated relevance to the current query, with earlier positions indicating higher relevance.
The role of this stage is to reduce a massive external capability space into a manageable ranked list of candidate skills that may be useful for the current task.

\paragraph{Skill Incorporation.}
Given the retrieved candidates $\mathcal{L}_k$, the agent must then determine whether any external skill should be used for the current task, and if so, which retrieved skills should be incorporated into the active problem-solving state and in what form. 
We denote this stage as
\begin{equation}
    \widetilde{\mathcal{S}} = G(q, \mathcal{L}_k; \mathcal{M}),
\end{equation}
where $\mathcal{M}$ is the underlying base language model and $\widetilde{\mathcal{S}}$ denotes the skill representations that are actually prepared and made available for downstream task solving. 
These representations may correspond to a selected subset of retrieved skills, or to transformed variants derived from them, such as rewritten, compressed, restructured, or model-adapted forms. 
Accordingly, skill incorporation is broader than merely selecting a subset from the retrieved list: it concerns whether and how external capabilities are converted into a form that the agent can subsequently use. 
Importantly, $\widetilde{\mathcal{S}}$ may be empty, either because the agent determines that its parametric capability is sufficient for the current task, or because none of the retrieved skills is deemed sufficiently relevant or useful to incorporate.

\paragraph{Skill Application.}
Finally, conditioned on the incorporated skills $\widetilde{\mathcal{S}}$, the agent applies the corresponding capabilities during task solving and produces a final response:
\begin{equation}
\hat{A} = F(q, \widetilde{\mathcal{S}}; \mathcal{M}).
\end{equation}
Importantly, successful incorporation does not guarantee successful application: the relevant skill may be present in the agent's active problem-solving state yet still fail to improve performance if the model cannot properly operationalize it during downstream task solving. 
This stage, therefore, assesses whether the incorporated skills are leveraged to improve downstream behavior, including whether the agent follows them correctly, invokes them at the right time, integrates them into its reasoning process, and adapts its behavior accordingly.

\section{Benchmark Construction}

SRA-Bench is built from three components:
(1) a set of capability-intensive \textbf{test instances},
(2) a set of manually constructed \textbf{gold skills} and corresponding instance-level annotations that associate each benchmark instance with one or more relevant skills, and
(3) a large external \textbf{skill corpus} formed by inserting these gold skills into a noisy web-collected skill collection containing realistic distractors.
This design makes the SRA setting empirically tractable by transforming skill augmentation into a decomposable evaluation problem: a system must retrieve the correct capability from a large, noisy skill corpus, correctly incorporate it, and ultimately use it to improve downstream task performance. 
We introduce the construction process in three subsections: 
{source dataset selection and test instance curation} (\S\ref{sec:test_instance_curation}), 
{gold skill construction} (\S\ref{sec:golden_skill_construction}), and 
{large-scale skill corpus collection} (\S\ref{sec:skill_corpus_assembly}).

\begin{table}[t]
\centering
\small
\caption{Overview of the six source datasets used to construct SRA-Bench. 
\textbf{Skill Mapping} indicates whether each benchmark instance is associated with one gold skill or multiple gold skills.}
\label{tab:datasets}
\resizebox{\textwidth}{!}{
\begin{tabular}{llrrcl}
\toprule
\textbf{Dataset} & \textbf{Capability Type} & \textbf{\#Inst.} & \textbf{\#Skills} & \textbf{Skill Mapping} & \textbf{Evaluation} \\
\midrule
\textsc{TheoremQA}~\cite{chen2023theoremqa} 
    & Theorem Application         & 747   & 320 & Single & Rule-Based \\
\textsc{LogicBench}~\cite{parmar2024logicbench} 
    & Logical Reasoning Patterns  & 760   & 19  & Single & Rule-Based \\
\textsc{ToolQA}~\cite{zhuang2023toolqa} 
    & Tool-Use Workflows          & 1,430 & 14  & Single & Rule-Based \\
\textsc{MedCalc-Bench}~\cite{khandekar2024medcalc} 
    & Medical Calculators         & 1,100 & 55  & Single & Rule-Based \\
\textsc{CHAMP}~\cite{mao2024champ} 
    & Mathematical Concepts       & 223   & 89  & Multi  & Rule-Based \\
\textsc{BigCodeBench}~\cite{zhuo2024bigcodebench} 
    & Software Libraries          & 1,140 & 139 & Multi  & Execution \\
\bottomrule
\end{tabular}
}
\end{table}

\subsection{Source Dataset Selection and Test Instance Curation}
\label{sec:test_instance_curation}

We begin the construction of SRA-Bench from the {instance side}: a collection of test problems that genuinely require reusable external capabilities rather than mere factual question answering (QA). 
To this end, we curate test instances from six existing benchmarks:
\textsc{TheoremQA}~\cite{chen2023theoremqa}, 
\textsc{LogicBench}~\cite{parmar2024logicbench}, 
\textsc{ToolQA}~\cite{zhuang2023toolqa}, 
\textsc{MedCalc-Bench}~\cite{khandekar2024medcalc}, 
\textsc{CHAMP}~\cite{mao2024champ}, 
and \textsc{BigCodeBench}~\cite{zhuo2024bigcodebench}. 
Together, these datasets cover diverse forms of capability use, including theorem-based reasoning, formal logic, tool-interleaved question answering, medical calculation, competition mathematics, and code generation. 
The resulting benchmark contains 5,400 test instances associated with 636 unique gold skills, as summarized in Table~\ref{tab:datasets}. 
Rather than starting from a pre-existing skill repository and searching for compatible tasks, we construct SRA-Bench in reverse: we first identify capability-intensive problem instances and then manually annotate the reusable skill(s) associated with them using structured signals from the source datasets.

Since existing benchmarks do not directly provide reusable skill annotations, the selection of source datasets is critical. 
We therefore select candidate datasets according to three criteria. 
First, the dataset should provide structured signals that reveal shared problem-solving patterns across multiple instances, enabling human annotators to reliably infer reusable capabilities. 
Second, the associated capability should encode \emph{how} to solve a class of problems, rather than directly revealing the answer to a particular instance. 
Third, the final task evaluation should remain objective and reproducible, so that downstream performance can be measured reliably. 
Together, these criteria ensure that the source datasets not only contain capability-intensive problems but also provide the annotation basis needed for constructing meaningful gold-skill supervision for SRA.

Under this design, each selected source instance is transformed into an SRA-Bench example consisting of a \textbf{user query}, a \textbf{ground-truth answer}, and one or more \textbf{gold skills}. 
Since reusable skill annotations are not available in existing benchmarks, we construct them manually using structured signals provided by the source datasets. 
For example, theorem names in \textsc{TheoremQA}, logic patterns in \textsc{LogicBench}, calculator types in \textsc{MedCalc-Bench}, and library names in \textsc{BigCodeBench} expose recurring capability structure across instances. 
These dataset-specific signals do not constitute gold skills themselves; instead, they guide annotators in abstracting instance-level problem-solving patterns into standardized reusable skills. 
Section~\ref{sec:golden_skill_construction} details this annotation process.

\subsection{Gold Skill Construction}
\label{sec:golden_skill_construction}

The source datasets provide structured annotations that indicate the capabilities involved in each instance, but they do not directly provide reusable skills. 
For example, annotations such as theorem names, logic patterns, calculator names, concept IDs, or library names identify \emph{what} capability is relevant, but they do not by themselves tell an agent \emph{when} to use that capability or \emph{how} to apply it for downstream tasks. 
Our goal in this stage is therefore to transform these source-side signals into explicit \textbf{gold skills}: reusable skill artifacts that can be retrieved and used under the SRA setting. 
Concretely, we construct one gold skill for each annotation category in the source datasets. 
Depending on the dataset, a gold skill may correspond to a theorem, a reasoning pattern, a tool-use workflow, a medical calculator, a mathematical concept, or a software library. 
Although these capability types differ in form, they can all be represented in the same way: as standalone skill artifacts that describe the applicability conditions and problem-solving procedure for a recurring class of tasks. 
Each benchmark instance is then associated with one or more gold skills according to its source annotation(s), forming the ground-truth skill supervision used in SRA-Bench. 
Since the available supervision signals differ markedly across source benchmarks, we defer full dataset-specific construction details and examples to Appendix~\ref{sec:appendix-construction}.

We build each gold skill through a two-stage process: LLM drafting followed by expert revision. 
For each annotation category, we first collect the available source materials, including the source dataset definition or description, representative instances annotated with this category, and external references when needed.
We then provide these materials to an LLM to generate an initial draft skill.
This draft serves only as a starting point.
It is then manually revised into the final gold skill, since draft skills often remain too tied to specific examples, miss important conditions or edge cases, or contain factual and procedural errors.
Across datasets, this revision is guided by three shared principles.
At the same time, different source benchmarks exhibit different recurring weaknesses in LLM drafts and therefore require additional dataset-specific refinements, which we detail in Appendix~\ref{sec:appendix-construction}. These shared principles are as follows.
First, {generality}: the final skill should describe a reusable method rather than a benchmark-specific template or paraphrase of example instances.
Second, {correctness}: formulas, reasoning procedures, tool usage, and executable components must be checked and revised against reliable references.
Third, {leakage control}: the skill should not reveal benchmark answers or encode shortcuts that trivialize evaluation.
To enforce this, overlapping examples are replaced with newly constructed ones, benchmark-specific constants are removed, and the final skill is written to preserve a clear application gap: even after retrieving the correct skill, the agent must still interpret the query, identify the appropriate case, extract instance-specific variables, and correctly execute the procedure.

The finalized gold skills are stored as standardized Markdown artifacts with a skill name, a short description, and procedural content that describes what the capability is, when it applies, how it should be used, and which common pitfalls to avoid. 
For capability types that inherently require execution, we additionally attach runnable resources such as Python implementations of medical calculators. 
This representation makes gold skills not merely annotation labels, but realistic skill artifacts that can be mixed into a large external corpus.

\subsection{Skill Corpus Collection}
\label{sec:skill_corpus_assembly}

To evaluate SRA under realistic large-scale conditions, gold skills must be retrieved from a noisy external corpus rather than presented in isolation. 
We therefore mix the 636 gold skills into a larger collection of 25,626 publicly available skill documents collected from the web, covering domains such as programming, data science, system administration, and general productivity. 
This yields a final skill corpus of 26,262 skills, of which only 2.4\% are gold.
To construct this background corpus, we follow a public-ecosystem crawling pipeline. 
Specifically, we collect publicly available skills from open web sources and community skill repositories, such as GitHub, Skills.sh, and the Hugging Face Hub. 
For each entry, we retrieve its associated skill documents or repository contents, and retain only those that expose sufficiently self-contained skill descriptions for standalone use.
We further remove inaccessible, malformed, or duplicate entries and normalize the remaining skills into a unified document format for indexing and retrieval. 
The resulting corpus is intended to approximate a realistic skill ecosystem in which high-value skills are sparse, heterogeneous, and mixed with many irrelevant or weakly related candidates.

\section{Study Design and Experimental Setup}

In this section, we present the study design of our systematic empirical study of scalable skill augmentation.
Rather than treating SRA as merely a new benchmark setting, we aim to understand the empirical factors that govern whether retrieved skills can actually improve agent performance.
We first present our study design and research questions, and then describe the experimental setup, including benchmarks, baselines, metrics, and implementation details.

\subsection{Study Design}
\label{sec:study-design}

The Skill Retrieval Augmentation (SRA) paradigm introduces a multi-stage pipeline in which an agent must retrieve, incorporate, and apply external skills to solve tasks that may exceed its native parametric capabilities. While the formulation is conceptually clean, each stage introduces distinct challenges and potential failure modes, and the interactions among stages further compound the difficulty of building effective SR-Agents. Therefore, before presenting experimental results, we first analyze the structure of the SRA problem and identify the key questions that must be answered to understand where the pipeline succeeds or fails.

At the highest level, the first question is whether SRA is useful at all. If retrieving and injecting external skills does not improve performance over skill-free baselines, then the practical value of the paradigm would be fundamentally limited. At the same time, even if SRA is beneficial in principle, real retrieval is inevitably noisy: retrieved candidate sets will contain distractors, partially relevant skills, and misleading entries in addition to useful ones. This means that the viability of SRA depends not only on whether skill augmentation helps, but also on whether SR-Agents remain effective under noisy retrieval conditions. Therefore, we first focus on the following two questions:
\vspace{0.3em}

\noindent\textbf{RQ1:} {Does the SRA paradigm improve agent performance over skill-free baselines, and how do different SR-Agents configurations compare in terms of end-task effectiveness?}

\noindent\textbf{RQ2:} {How robust are current SR-Agents to retrieval noise? Specifically, can they still identify and effectively utilize relevant skills when candidate sets contain irrelevant distractors, and how does robustness vary across different SR-Agents designs?}
\vspace{0.3em}

The next question concerns the retrieval stage itself. 
Since skills are heterogeneous capability packages rather than text passages that primarily convey factual world knowledge, it is unclear whether existing retrieval methods transfer effectively to this setting, or whether lexical and dense retrievers exhibit different strengths for skill retrieval. 
Moreover, retrieval quality alone does not fully determine end-task success: SRA is a coupled pipeline, and gains from better retrieval may be amplified by downstream incorporation and application. 
Therefore, beyond evaluating retrieval in isolation, we must also ask how strongly retrieval quality actually affects end-to-end performance. 
Accordingly, we study the following two questions:
\vspace{0.3em}

\noindent\textbf{RQ3:} {How effective are existing retrieval methods at identifying relevant skills from large-scale skill corpus, and how do classical lexical matching approaches compare with dense retrieval methods in the skill retrieval setting?}

\noindent\textbf{RQ4:} {To what extent does retrieval quality influence end-to-end SR-Agents performance? Does better retrieval consistently lead to better task outcomes, or are the gains mediated or attenuated by downstream incorporation and application stages?}
\vspace{0.3em}

Finally, even strong retrieval does not guarantee effective skill use, because the language model must properly incorporate the retrieved candidates. This raises two closely related questions about skill-loading behavior. First, a rational agent should be \emph{relevance-aware}: it should be more likely to load skills when the retrieved candidate set actually contains a relevant skill. Second, a rational agent should be \emph{need-aware}: it should preferentially load external skills for tasks that exceed its native capability, while refraining from unnecessary loading for tasks it can already solve on its own. These two properties determine whether retrieval success translates into effective utilization. Therefore, we finally focus on the following two questions:
\vspace{0.3em}

\noindent\textbf{RQ5:} {During skill incorporation, can current LLMs distinguish between retrieved candidate sets that contain a relevant (gold) skill and those that do not? Does the presence of a gold skill among retrieved candidates meaningfully influence the agent's skill-loading behavior?}

\noindent\textbf{RQ6:} {Do current LLMs exhibit need-aware skill-loading behavior during incorporation? Specifically, are agents more inclined to load external skills for tasks that exceed their native capabilities than for tasks they can already solve without external augmentation?}
\vspace{0.3em}

Taken together, these six questions cover the major uncertainty sources of the SRA pipeline: whether the paradigm is beneficial at all (RQ1), whether it remains effective under noisy retrieval conditions (RQ2), how well skill retrieval itself can be solved (RQ3), whether retrieval improvements translate into downstream gains (RQ4), and whether current LLMs incorporate skills in a relevance-aware and need-aware manner (RQ5--RQ6). By answering these questions systematically, we aim to provide a more diagnostic understanding of scalable skill augmentation and to identify the key bottlenecks that currently limit SR-Agents' performance.

\subsection{Experimental Setup}
\label{sec:experimental-setup}

To provide a consistent basis for answering the research questions in \S\ref{sec:study-design}, we standardize the main experimental components shared across our study, including benchmarks, evaluated models, skill-use strategies, and evaluation metrics. Since this paper presents a paradigm and systematic study rather than a single method under a single protocol, experiment-specific variations are deferred to the corresponding subsections.

\paragraph{Benchmarks and Skill Corpus.}
We conduct experiments on SRA-Bench, which comprises six capability-intensive benchmarks: TheoremQA~\cite{chen2023theoremqa}, LogicBench~\cite{parmar2024logicbench}, ToolQA~\cite{zhuang2023toolqa}, CHAMP~\cite{mao2024champ}, MedCalc-Bench~\cite{khandekar2024medcalc}, and BigCodeBench~\cite{zhuo2024bigcodebench}. These datasets cover diverse task settings, including mathematical reasoning, formal logic, tool use, and code generation. Each instance is paired with annotated gold skill(s) and evaluated against a shared external skill corpus, enabling decomposed analysis of retrieval, incorporation, and end-task execution.

\paragraph{Evaluation Metrics.}
We evaluate the SRA pipeline from both retrieval and end-task perspectives. For skill retrieval, we report Recall@$K$ and nDCG@$K$. For end-task performance, each benchmark follows its standard evaluation protocol: rule-based exact match or accuracy for reasoning and QA datasets, and pass@1 based on unit-test execution for BigCodeBench.

\paragraph{Selected Models.}
We evaluate six LLMs from three open-weight families: Qwen3-4B, Qwen3-32B, and Qwen3-235B-A22B~\cite{yang2025qwen3}; Llama-3.1-8B-Instruct~\cite{grattafiori2024llama} and Llama-3.3-70B-Instruct~\cite{meta2024llama33modelcard}; and Mistral-Small-3.1-24B-Instruct-2503~\cite{mistral2025small31}. For behavioral analyses of skill-loading dynamics (RQ5 and RQ6), we also include proprietary frontier models GLM-5.1~\cite{zai2026glm51blog} and GPT-5.4~\cite{openai2026gpt54}. All models are served with a 128K-token context window and sampling temperature 0.7.

\paragraph{Skill-Use Strategies.}
We study both non-augmented and skill-augmented inference settings. The non-augmented baseline is \textbf{LLM Direct}, where the model solves the task using only its parametric knowledge. The upper-bound setting is \textbf{Oracle Skill}, where the annotated gold skill(s) are directly provided to the model. For skill-retrieval-augmented inference, we consider three representative strategies. 
\textbf{Full-Skill Injection} injects the complete content of the top-$k$ retrieved skills into the task context. 
This represents the most direct use of retrieved skills, but also exposes the model to all retrieved content, including potentially irrelevant skills. 
\textbf{LLM Selection} instead exposes only metadata for the retrieved candidates, asks the model to select the single most relevant skill, and then injects the full content of the selected skill. 
\textbf{Progressive Disclosure} follows an OpenClaw-style design, where the model is given a compact skill catalog and can selectively load full skill content on demand during inference.
The exact prompt templates used for these skill-use baselines are provided in Appendix~\ref{sec:appendix-inference}.

\paragraph{Implementation Details.}
When a skill is injected, its full content is prepended before the original task prompt. In selection-based settings, the model is given a list of candidates, each with a skill name and description, and must select the most relevant one before generating an answer. In the Progressive Disclosure setting, the model interacts with a compact skill catalog and may explicitly issue skill-loading actions to inspect a skill's full content during reasoning. 
% The detailed interaction protocol and candidate construction procedure are specified in the corresponding experiment sections.
In retrieval-based experiments, candidate skills are retrieved from the shared external skill corpus in response to the input query. BM25 serves as the default retriever in the main experiments, while additional sparse, dense, hybrid, and reranking-based retrieval variants are introduced in \S ~\ref{sec:rq3-retrieval}.

\section{Systematic Empirical Study}
\label{sec:main-results}

\subsection{RQ1: Does Skill Retrieval Augmentation Improve Agent Performance?}
\label{sec:RQ1}

\begin{table*}[t]
\caption{End-task performance (\%) of different models and skill-use methods on SRA-Bench. Oracle is included for comparison. The average is computed over all instances. For retrieval-based methods, Full-Skill Injection uses the top-1 skill retrieved by BM25, while LLM Selection and Progressive Disclosure select skills from the top-50 BM25 candidates. 
The best results are in bold, and the second-best results are underlined. Statistical significance is assessed using a paired t-test, where $^{*}$ indicates performance significantly worse than the bold method at $p < 0.05$. }
\label{tab:main_results_all_models}
\vspace{1.5mm}
\centering
\small
\setlength{\tabcolsep}{5pt}
\renewcommand{\arraystretch}{1.15}
\resizebox{\textwidth}{!}{
\begin{tabular}{llccccccc}
\toprule
Model & Method & TheoremQA & LogicBench & ToolQA & CHAMP & MedCalc & BigCodeBench & Average \\
\midrule

\multirow{5}{*}{Llama-3.1-8B}
& LLM Direct               & 32.4$^{*}$ & 54.6$^{*}$ & 16.7$^{*}$ & 22.4$^{*}$ & 26.9$^{*}$ & 32.3\hphantom{$^{*}$} & 29.8$^{*}$ \\
& Oracle Skill             & \textbf{49.4}\hphantom{$^{*}$} & \textbf{69.5}\hphantom{$^{*}$} & \textbf{23.3}\hphantom{$^{*}$} & \textbf{40.8}\hphantom{$^{*}$} & \textbf{62.0}\hphantom{$^{*}$} & \textbf{35.2}\hphantom{$^{*}$} & \textbf{44.5}\hphantom{$^{*}$} \\
& Full-Skill Injection                & 36.5$^{*}$ & \underline{58.3}$^{*}$ & 13.6$^{*}$ & \underline{27.8}$^{*}$ & 36.7$^{*}$ & \underline{34.2}\hphantom{$^{*}$} & 32.7$^{*}$ \\
& LLM Selection               & 35.1$^{*}$ & 53.8$^{*}$ & \underline{19.4}$^{*}$ & 24.7$^{*}$ & 57.0$^{*}$ & 32.1$^{*}$ & \underline{37.0}$^{*}$ \\
& Progressive Disclosure    & \underline{36.9}$^{*}$ & 50.0$^{*}$ & 16.4$^{*}$ & 25.1$^{*}$ & \underline{59.6}\hphantom{$^{*}$} & 31.4$^{*}$ & 36.3$^{*}$ \\
\midrule

\multirow{5}{*}{Llama-3.3-70B}
& LLM Direct               & 59.6$^{*}$ & 60.5$^{*}$ & 30.9$^{*}$ & 56.5$^{*}$ & 53.9$^{*}$ & 45.0$^{*}$ & 47.8$^{*}$ \\
& Oracle Skill             & \textbf{68.5}\hphantom{$^{*}$} & \textbf{81.1}\hphantom{$^{*}$} & \textbf{48.7}\hphantom{$^{*}$} & \textbf{68.6}\hphantom{$^{*}$} & \textbf{79.7}\hphantom{$^{*}$} & \textbf{54.7}\hphantom{$^{*}$} & \textbf{64.4}\hphantom{$^{*}$} \\
& Full-Skill Injection                & 60.1$^{*}$ & 66.1$^{*}$ & 35.0$^{*}$ & \underline{61.9}$^{*}$ & 59.5$^{*}$ & \underline{52.9}\hphantom{$^{*}$} & 52.7$^{*}$ \\
& LLM Selection               & \underline{62.7}$^{*}$ & \underline{69.7}$^{*}$ & \underline{43.3}$^{*}$ & 58.7$^{*}$ & \underline{79.4}\hphantom{$^{*}$} & 50.4$^{*}$ & \underline{59.2}$^{*}$ \\
& Progressive Disclosure    & 53.3$^{*}$ & 65.3$^{*}$ & 29.9$^{*}$ & 54.3$^{*}$ & 59.5$^{*}$ & 45.7$^{*}$ & 48.5$^{*}$ \\
\midrule

\multirow{5}{*}{Mistral3.1-24B}
& LLM Direct               & 49.3$^{*}$ & 56.1$^{*}$ & 29.2$^{*}$ & 52.5$^{*}$ & 49.6$^{*}$ & 41.1$^{*}$ & 43.4$^{*}$ \\
& Oracle Skill             & \textbf{66.3}\hphantom{$^{*}$} & \textbf{78.6}\hphantom{$^{*}$} & \textbf{48.2}\hphantom{$^{*}$} & \textbf{66.8}\hphantom{$^{*}$} & \textbf{78.6}\hphantom{$^{*}$} & \textbf{54.0}\hphantom{$^{*}$} & \textbf{63.2}\hphantom{$^{*}$} \\
& Full-Skill Injection                & 58.8$^{*}$ & 57.2$^{*}$ & 34.4$^{*}$ & \underline{59.6}$^{*}$ & 54.6$^{*}$ & 45.6$^{*}$ & 48.5$^{*}$ \\
& LLM Selection               & \underline{59.0}$^{*}$ & \underline{66.7}$^{*}$ & \underline{42.6}$^{*}$ & 55.6$^{*}$ & \underline{76.3}$^{*}$ & \underline{46.8}$^{*}$ & \underline{56.6}$^{*}$ \\
& Progressive Disclosure    & 54.8$^{*}$ & 55.8$^{*}$ & 27.1$^{*}$ & 54.7$^{*}$ & 62.6$^{*}$ & 42.8$^{*}$ & 46.7$^{*}$ \\
\midrule

\multirow{5}{*}{Qwen3-235B}
& LLM Direct               & 61.6$^{*}$ & 76.1$^{*}$ & 36.4$^{*}$ & 66.4$^{*}$ & 58.2$^{*}$ & 46.7$^{*}$ & 53.3$^{*}$ \\
& Oracle Skill             & 65.9\hphantom{$^{*}$} & \textbf{86.6}\hphantom{$^{*}$} & \textbf{52.3}\hphantom{$^{*}$} & \textbf{75.8}\hphantom{$^{*}$} & \textbf{84.5}\hphantom{$^{*}$} & \textbf{57.0}\hphantom{$^{*}$} & \textbf{67.5}\hphantom{$^{*}$} \\
& Full-Skill Injection                & 64.4$^{*}$ & 72.8$^{*}$ & 38.0$^{*}$ & 68.6$^{*}$ & 66.2$^{*}$ & \underline{50.4}$^{*}$ & 56.2$^{*}$ \\
& LLM Selection               & \underline{66.8}\hphantom{$^{*}$} & 79.5$^{*}$ & \underline{45.4}$^{*}$ & \underline{73.5}\hphantom{$^{*}$} & \underline{82.5}\hphantom{$^{*}$} & 49.7$^{*}$ & \underline{62.8}$^{*}$ \\
& Progressive Disclosure    & \textbf{68.1}\hphantom{$^{*}$} & \underline{81.1}$^{*}$ & 36.9$^{*}$ & 72.2\hphantom{$^{*}$} & 77.1$^{*}$ & 50.3$^{*}$ & 59.9$^{*}$ \\
\midrule

\multirow{5}{*}{Qwen3-32B}
& LLM Direct               & 57.4$^{*}$ & 75.9$^{*}$ & 35.0$^{*}$ & 65.5\hphantom{$^{*}$} & 53.9$^{*}$ & 43.9$^{*}$ & 50.8$^{*}$ \\
& Oracle Skill             & \textbf{71.6}\hphantom{$^{*}$} & \textbf{86.6}\hphantom{$^{*}$} & \textbf{51.2}\hphantom{$^{*}$} & \textbf{70.9}\hphantom{$^{*}$} & \textbf{83.5}\hphantom{$^{*}$} & \textbf{55.2}\hphantom{$^{*}$} & \textbf{67.2}\hphantom{$^{*}$} \\
& Full-Skill Injection                & 68.1$^{*}$ & 70.5$^{*}$ & 36.2$^{*}$ & \underline{70.4}\hphantom{$^{*}$} & 59.5$^{*}$ & \underline{49.0}$^{*}$ & 54.3$^{*}$ \\
& LLM Selection               & \underline{69.3}\hphantom{$^{*}$} & \underline{81.1}$^{*}$ & \underline{44.1}$^{*}$ & 65.9\hphantom{$^{*}$} & \underline{82.5}\hphantom{$^{*}$} & 48.1$^{*}$ & \underline{62.4}$^{*}$ \\
& Progressive Disclosure    & 64.9$^{*}$ & 74.7$^{*}$ & 35.5$^{*}$ & 58.7$^{*}$ & 71.1$^{*}$ & 44.7$^{*}$ & 55.3$^{*}$ \\
\midrule

\multirow{5}{*}{Qwen3-4B}
& LLM Direct               & 50.7$^{*}$ & \underline{75.0}$^{*}$ & 25.6$^{*}$ & 56.1$^{*}$ & 22.0$^{*}$ & 36.4$^{*}$ & 38.8$^{*}$ \\
& Oracle Skill             & \textbf{69.7}\hphantom{$^{*}$} & \textbf{85.1}\hphantom{$^{*}$} & \textbf{47.1}\hphantom{$^{*}$} & \textbf{70.9}\hphantom{$^{*}$} & \textbf{73.5}\hphantom{$^{*}$} & \textbf{45.5}\hphantom{$^{*}$} & \textbf{61.6}\hphantom{$^{*}$} \\
& Full-Skill Injection                & \underline{64.0}$^{*}$ & 69.1$^{*}$ & 30.2$^{*}$ & \underline{68.6}\hphantom{$^{*}$} & 36.1$^{*}$ & \underline{41.5}$^{*}$ & 45.5$^{*}$ \\
& LLM Selection               & 63.1$^{*}$ & 70.0$^{*}$ & \underline{39.0}$^{*}$ & 62.3$^{*}$ & \underline{65.7}$^{*}$ & 39.7$^{*}$ & \underline{53.3}$^{*}$ \\
& Progressive Disclosure    & 52.7$^{*}$ & 65.7$^{*}$ & 26.7$^{*}$ & 60.1$^{*}$ & 45.0$^{*}$ & 37.7$^{*}$ & 43.2$^{*}$ \\
\bottomrule
\end{tabular}
}
\end{table*}

We begin with the most fundamental question for the SRA paradigm: does retrieving and injecting external skills actually improve an agent's performance compared to solving tasks directly from parametric knowledge alone? 
Table~\ref{tab:main_results_all_models} reports end-task results across six benchmarks, six LLMs, and five skill-use settings, which differ in how external skills are accessed and incorporated during inference. 
At the two extremes, \textit{LLM Direct} measures the model's native task-solving ability without any external skill support. In contrast, \textit{Oracle Skill} directly provides the annotated gold skill(s), serving as an upper bound on the potential value of correct skill access. 
Between these two extremes, we evaluate three practical SRA settings. 
\textit{Full-Skill Injection} directly injects the top-1 skill retrieved by BM25 into the model context.
\textit{LLM Selection} first retrieves a larger candidate pool (top-50) and then requires the model to choose one skill to load.
\textit{Progressive Disclosure} follows an OpenClaw-style protocol, where the agent is shown only a compact catalog of retrieved candidates and may decide for itself whether to reveal and load a skill on demand.
Importantly, in both LLM Selection and Progressive Disclosure, the initial context contains only the name and description of each candidate among the retrieved top-50 skills, rather than the full skill content; the full skill is exposed only after a skill is explicitly selected.
Conceptually, Progressive Disclosure is the most rational setting for an agentic system, as an ideal agent should only invoke external capabilities when its native parametric knowledge falls short, rather than injecting skills regardless of necessity.

Several conclusions emerge clearly from the results. 
First, the answer to RQ1 is unambiguously positive: external skills can substantially improve agent performance. 
When the correct skill is provided, \textit{Oracle Skill} consistently outperforms \textit{LLM Direct} across essentially all model-benchmark combinations, often by a large margin. 
This confirms a central premise of SRA-Bench: many tasks in our benchmark genuinely benefit from externalized procedural knowledge, and such skills are not redundant auxiliary context. 
When correctly accessed and incorporated, they materially extend what agents can accomplish beyond their native parametric capability.

Second, practical retrieval-based SRA methods can already yield meaningful gains, but these gains are far from uniform or guaranteed. 
In many model-benchmark combinations, \textit{Full-Skill Injection} and especially \textit{LLM Selection} improve over \textit{LLM Direct}, showing that the value of skill augmentation is not confined to idealized oracle settings.
However, these improvements are highly uneven across models, tasks, and exposure strategies, and retrieval-based methods remain substantially below the \textit{Oracle Skill} upper bound in most cases. 
In other words, our results clearly establish the {potential} of skill retrieval augmentation, but also show that realizing this potential robustly under practical inference settings remains an open challenge.

Third, and most importantly, the effectiveness of SRA depends not only on whether the right skill is retrieved, but also on how the retrieved skill is exposed to the agent. 
Among the practical methods, LLM Selection provides the strongest overall trade-off between effectiveness and stability, consistently translating retrieval results into downstream gains and, in many cases, substantially narrowing the gap to the oracle upper bound.
This suggests that, under current models, selection-based skill exposure is a more reliable way to operationalize external capabilities than simply injecting the top-ranked skill or leaving the loading decision entirely to the agent. By contrast, Progressive Disclosure exhibits a much less stable pattern: while it can improve performance in some settings, its gains are inconsistent, and it rarely matches the strongest selection-based configurations. The implication is that the bottleneck in SRA is not retrieval alone but controlled skill utilization: scalable skill augmentation requires not only access to a large skill corpus but also a reliable mechanism for deciding when and how retrieved skills should enter the reasoning process.

Overall, RQ1 supports a positive but qualified conclusion: Skill Retrieval Augmentation can indeed improve agent performance across a diverse range of models and tasks.
Yet these benefits are not guaranteed by retrieval alone: they also depend critically on how retrieved skills are presented, selected, and integrated into the agent’s reasoning process.
This observation directly motivates RQ2: once retrieval becomes noisy and candidate sets contain irrelevant distractors, can current SR-Agents still identify and utilize the right skills effectively?

\subsection{RQ2: How Robust Are SR-Agents to Retrieval Noise?}
\label{sec:RQ2}

\begin{figure*}[t]
\centering
    \includegraphics[width=\textwidth]{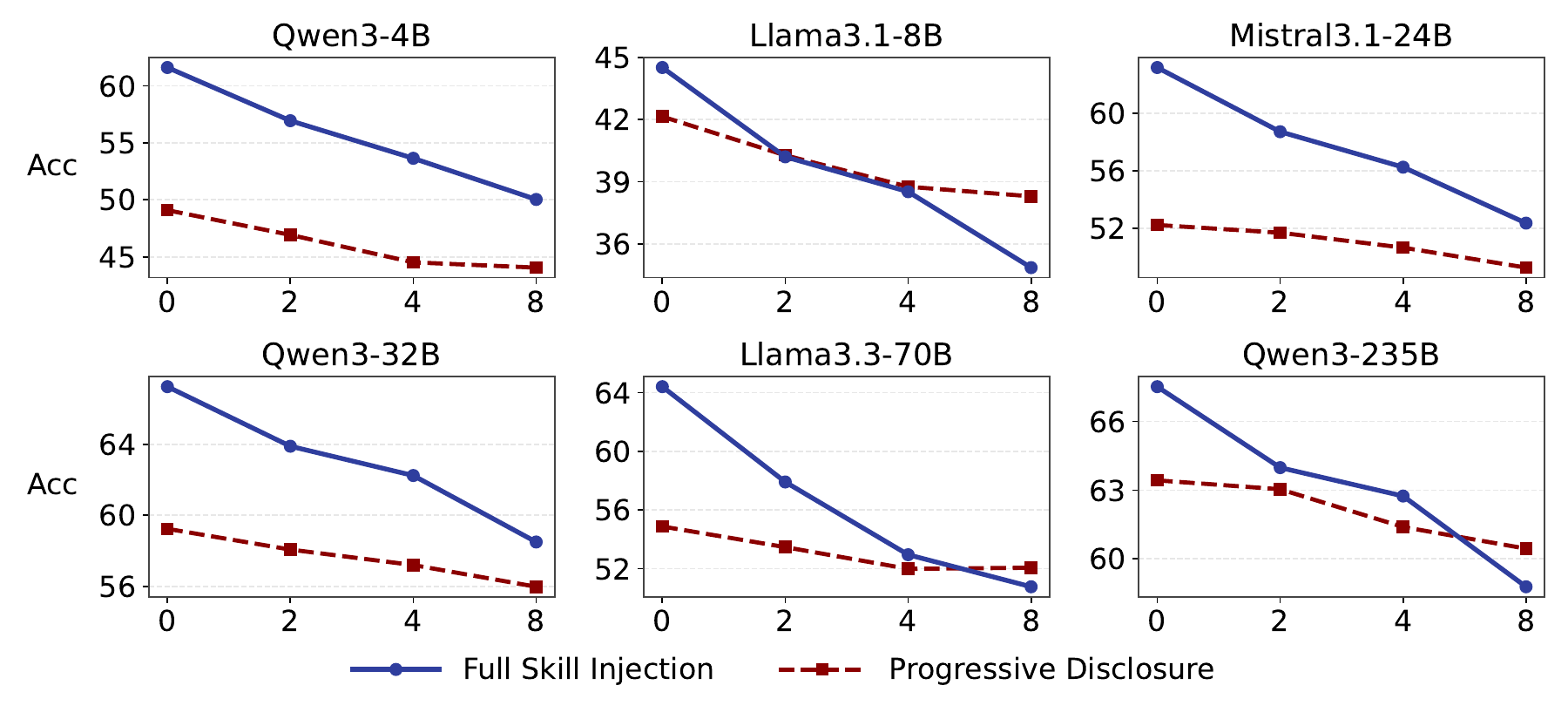}
    \caption{End-task accuracy as the number of hard-negative distractor skills increases, with the gold skill always included. Full Skill Injection is consistently more brittle to noise, while \textit{Progressive Disclosure} remains more robust across models.} 
    \label{fig:rq2_noise}
\end{figure*}

\begin{figure*}[t]
\centering
    \includegraphics[width=\textwidth]{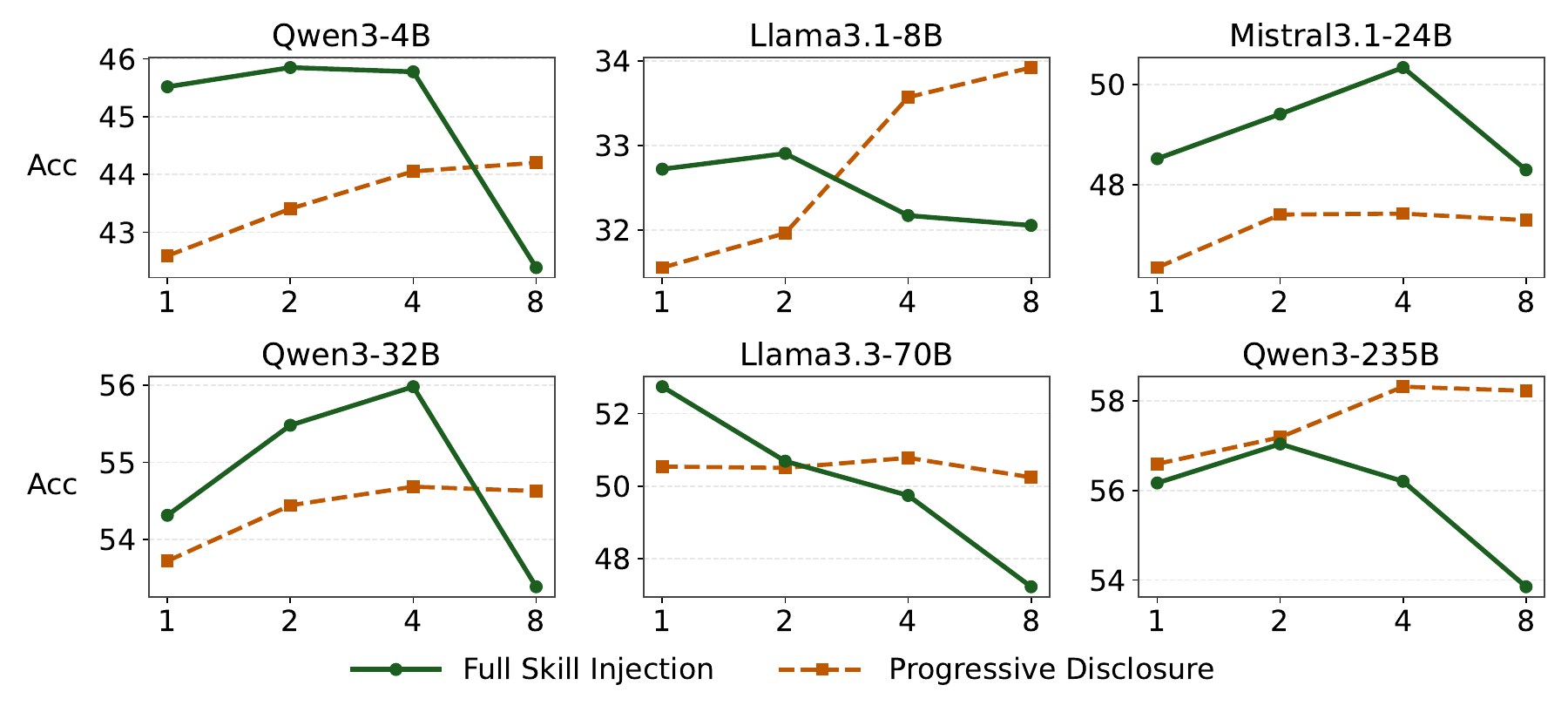}
    \caption{End-task accuracy under BM25 top-$k$ skill exposure. Increasing $k$ may improve relevant-skill coverage, but \textit{Full Skill Injection} typically improves only at small $k$ and then degrades as more full skill content is injected, suggesting inter-skill interference. In contrast, \textit{Progressive Disclosure} remains more stable by exposing compact metadata and loading full skill content selectively.}
    \label{fig:rq2_topk}
\end{figure*}

\subsubsection{Robustness under Controlled Distractors}
The practical effectiveness of SRA depends not only on whether the correct skill can be retrieved, but also on whether the agent can still identify and use it once the candidate set becomes noisy. In realistic deployments, retrieved results inevitably include irrelevant yet plausible distractors, which may interfere with skill selection, loading, and downstream task solving. To isolate this incorporation challenge from retrieval itself, we construct a controlled setting in which the gold skill is {always} included in the candidate set, together with $N$ hard-negative distractor skills, where $N \in \{0,2,4,8\}$. 
The distractors are constructed by alternately selecting non-gold, non-duplicate candidates from the BM25 and BGE retrieval lists in rank order, starting at rank 1, to capture both lexical and semantic distractor patterns. The final candidate order is then shuffled to avoid positional bias.
We evaluate two representative skill-exposure strategies: \textit{Full Skill Injection}, which injects the full content of all candidate skills into the prompt, and \textit{Progressive Disclosure}, which presents only the name and description, and injects the full skill content only after an explicit selection. Figure~\ref{fig:rq2_noise} reports the end-task performance under increasing distractor levels.

Figure~\ref{fig:rq2_noise} shows that current SR-Agents are highly brittle to retrieval noise, even when the correct skill is already present in the candidate set. 
Across models, adding hard-negative distractors consistently degrades end-task performance, indicating that recall alone is insufficient for effective skill augmentation: once multiple plausible but irrelevant candidates coexist with the correct skill, the primary failure mode shifts to incorporation, i.e., whether the agent can reliably identify, preserve, and utilize the truly useful skill under distraction. 
Notably, this brittleness varies substantially across skill-exposure strategies.
Under \textit{Full Skill Injection} setting, performance drops sharply as distractors accumulate, suggesting that directly injecting multiple full skills creates substantial interference from both prompt overload and procedural confusion. 
By contrast, \textit{Progressive Disclosure} is more stable: by withholding full skill content until an explicit reveal decision is made, it reduces irrelevant exposure and often matches or even surpasses Full Skill Injection under heavier noise. 
Notably, this robustness gap does not disappear with model scale. Larger models are not consistently better at suppressing distractors, and both small and strong models can fail once the candidate set becomes noisy. 
Taken together, these results show that scalable skill augmentation is not bottlenecked by retrieval alone, but by the agent's ability to selectively expose and incorporate retrieved skills under interference, making controlled skill exposure a core requirement for robust SR-Agents.

\subsubsection{Robustness under Retrieved Top-k Skill Exposure}

The controlled distractor setting isolates the incorporation problem by guaranteeing that the gold skill is present. In practical retrieval, however, increasing the number of exposed candidates introduces a more fundamental trade-off: a larger $k$ may improve the chance of covering relevant skills, but it also exposes the agent to more irrelevant or partially relevant skills. To examine this trade-off, we further conduct a BM25 top-$k$ sweep, where $k \in \{1,2,4,8\}$. Under \emph{Full Skill Injection}, the complete contents of the top-$k$ retrieved skills are inserted into the task context. Under \emph{Progressive Disclosure}, only compact metadata is initially exposed, and the agent may selectively load full skill content on demand. Figure~\ref{fig:rq2_topk} summarizes the average end-task performance under this setting.

Figure~\ref{fig:rq2_topk} reveals a clear divergence between the two exposure strategies. \emph{Full Skill Injection} often benefits from increasing $k$ only at the very beginning, suggesting that exposing a small number of additional retrieved skills can improve relevant-skill coverage. However, as $k$ continues to grow, its performance typically declines. This inverted trend shows that naively injecting more retrieved skills does not reliably convert higher retrieval coverage into better task performance. Instead, additional full skill contents can introduce inter-skill interference, distract the model from the truly useful capabilities, and perturb downstream reasoning. In contrast, \emph{Progressive Disclosure} remains much more stable as $k$ increases. By exposing only compact skill metadata before full-content loading, it allows the agent to inspect a broader candidate set without forcing all procedural details into the reasoning context. As a result, the comparison between the two strategies changes with candidate-set size: \emph{Full Skill Injection} can be stronger when only the top-1 skill is exposed, where little distractor noise is present and direct access to the full skill is useful; but under broader top-$k$ exposure, this advantage largely disappears and is often reversed. This finding explains why Full Skill Injection may appear competitive in single-skill settings, while Progressive Disclosure provides a more scalable design for realistic SR-Agents, where agents must choose among multiple retrieved candidates rather than blindly trust a single retrieved skill.

Together, the controlled distractor experiment and the BM25 top-$k$ sweep show that retrieval noise affects SRA through two complementary mechanisms. Even when the gold skill is guaranteed to be available, distractors can prevent the agent from incorporating it correctly; and when more retrieved skills are exposed to improve coverage, uncontrolled full-content injection can turn candidate-set expansion into procedural interference. Therefore, robustness in SRA is not merely a property of the retriever, but also a property of the skill-exposure mechanism. Controlled exposure and selective loading are thus central requirements for building robust SR-Agents at scale.

\subsection{RQ3: How Effective Are Existing Retrievers for Skill Retrieval?}
\label{sec:rq3-retrieval}

\begin{table*}[t]
\caption{Skill retrieval performance across different retrieval methods. We report Recall@1 and Recall@10 for each dataset. The upper block shows first-stage retrievers, while the lower block shows LLM-based rerankers applied to the top-50 candidates retrieved by BM25.}
\label{tab:retrieval_recall_merged}
\vspace{1.5mm}
\centering
\scriptsize
\setlength{\tabcolsep}{2pt}
\resizebox{\textwidth}{!}{
\begin{tabular}{lccccc ccccc ccc}
\toprule
\multirow{2}{*}{Method} 
& \multicolumn{2}{c}{TheoremQA} 
& \multicolumn{2}{c}{LogicBench} 
& \multicolumn{2}{c}{ToolQA} 
& \multicolumn{2}{c}{CHAMP} 
& \multicolumn{2}{c}{MedCalc-Bench} 
& \multicolumn{2}{c}{BigCodeBench} \\
\cmidrule(lr){2-3} \cmidrule(lr){4-5} \cmidrule(lr){6-7} \cmidrule(lr){8-9} \cmidrule(lr){10-11} \cmidrule(lr){12-13}
& R@1 & R@10 & R@1 & R@10 & R@1 & R@10 & R@1 & R@10 & R@1 & R@10 & R@1 & R@10 \\
\midrule
BM25 & 57.2 & 80.7 & 12.0 & 36.1 & 7.0 & 55.1 & 13.2 & 36.1 & 29.3 & 69.2 & 23.6 & 61.1 \\
TF-IDF & 41.4 & 68.8 & 1.8 & 18.2 & 7.0 & 35.0 & 7.2 & 25.7 & 37.5 & 71.4 & 20.9 & 60.2 \\
BGE & 66.8 & 86.1 & 4.1 & 20.5 & 32.2 & 83.4 & 9.8 & 34.0 & 41.4 & 70.1 & 20.7 & 62.1 \\
Contriever & 52.1 & 75.6 & 5.5 & 18.4 & 21.2 & 42.7 & 3.7 & 29.3 & 34.9 & 66.9 & 19.0 & 54.1 \\
Hybrid & 57.2 & 90.0 & 12.0 & 33.6 & 7.0 & 83.5 & 13.2 & 41.4 & 29.3 & 67.8 & 23.6 & 68.4 \\
\midrule
Llama-3.1-8B & 58.8 & 83.7 & 15.9 & 42.0 & 25.7 & 66.4 & 15.8 & 41.6 & 86.0 & 91.5 & 22.9 & 68.7 \\
Llama-3.3-70B & 76.0 & 88.6 & 27.4 & 55.4 & 51.0 & 76.9 & 22.5 & 47.8 & 89.5 & 92.5 & 27.4 & 78.6 \\
Mistral3.1-24B & 74.2 & 89.0 & 26.7 & 53.2 & 53.9 & 76.6 & 18.0 & 49.3 & 91.1 & 92.5 & 27.7 & 79.4 \\
Qwen3-235B & 75.4 & 88.8 & 30.9 & 56.4 & 56.4 & 76.2 & 22.1 & 50.2 & 92.3 & 92.5 & 27.2 & 80.0 \\
Qwen3-32B & 77.4 & 88.8 & 31.4 & 55.3 & 43.7 & 74.8 & 22.3 & 49.1 & 91.2 & 92.4 & 28.2 & 79.7 \\
Qwen3-4B & 69.5 & 87.1 & 21.8 & 43.3 & 39.9 & 70.8 & 18.5 & 44.0 & 85.9 & 91.2 & 26.3 & 72.5 \\
\bottomrule
\end{tabular}
}

\end{table*}

\begin{table*}[t]
\caption{Skill retrieval performance across different retrieval methods. We report N@1 (nDCG@1) and N@10(nDCG@10) for each dataset. The upper block shows first-stage retrievers, while the lower block shows LLM-based rerankers applied to the top-50 candidates retrieved by BM25.}
\label{tab:retrieval_N_merged}
\vspace{1.5mm}
\centering
\scriptsize
\setlength{\tabcolsep}{2pt}
\resizebox{\textwidth}{!}{
\begin{tabular}{lcccccccccccc}
\toprule
\multirow{2}{*}{Method} 
& \multicolumn{2}{c}{TheoremQA} 
& \multicolumn{2}{c}{LogicBench} 
& \multicolumn{2}{c}{ToolQA} 
& \multicolumn{2}{c}{CHAMP} 
& \multicolumn{2}{c}{MedCalc-Bench} 
& \multicolumn{2}{c}{BigCodeBench} \\
\cmidrule(lr){2-3} \cmidrule(lr){4-5} \cmidrule(lr){6-7} \cmidrule(lr){8-9} \cmidrule(lr){10-11} \cmidrule(lr){12-13}
& N@1 & N@10 & N@1 & N@10 & N@1 & N@10 & N@1 & N@10 & N@1 & N@10 & N@1 & N@10 \\
\midrule
BM25 & 57.2 & 69.2 & 12.0 & 22.6 & 7.0 & 27.0 & 20.6 & 27.2 & 29.3 & 44.7 & 61.7 & 55.4 \\
TF-IDF & 41.4 & 54.4 & 1.8 & 8.3 & 7.0 & 18.9 & 12.1 & 17.1 & 37.5 & 52.0 & 55.2 & 52.0 \\
BGE & 66.8 & 75.9 & 4.1 & 11.1 & 32.2 & 58.6 & 13.5 & 22.6 & 41.4 & 54.5 & 54.0 & 53.6 \\
Contriever & 52.1 & 63.8 & 5.5 & 10.9 & 21.2 & 31.9 & 5.4 & 16.4 & 34.9 & 49.4 & 49.2 & 47.9 \\
Hybrid & 57.2 & 75.3 & 12.0 & 21.0 & 7.0 & 44.5 & 20.6 & 28.6 & 29.3 & 47.4 & 61.7 & 59.9 \\
\midrule
Llama-3.1-8B & 58.8 & 71.5 & 15.9 & 28.8 & 25.7 & 47.5 & 25.6 & 32.1 & 86.0 & 88.9 & 59.2 & 61.6 \\
Llama-3.3-70B & 76.0 & 82.8 & 27.4 & 41.8 & 51.0 & 66.2 & 35.0 & 39.1 & 89.5 & 91.3 & 71.1 & 72.1 \\
Mistral3.1-24B & 74.2 & 81.9 & 26.7 & 40.3 & 53.9 & 67.3 & 27.4 & 37.0 & 91.1 & 91.9 & 72.0 & 72.9 \\
Qwen3-235B & 75.4 & 83.1 & 30.9 & 44.4 & 56.4 & 68.1 & 34.5 & 41.5 & 92.3 & 92.4 & 70.7 & 73.1 \\
Qwen3-32B & 77.4 & 83.5 & 31.4 & 44.2 & 43.7 & 62.7 & 35.4 & 41.0 & 91.2 & 91.9 & 73.2 & 74.1 \\
Qwen3-4B & 69.5 & 79.0 & 21.8 & 32.3 & 39.9 & 57.5 & 30.0 & 35.2 & 85.9 & 88.9 & 68.4 & 66.6 \\
\bottomrule
\end{tabular}
}

\end{table*}

We next examine whether existing retrieval methods can reliably identify relevant skills from a large external skill corpus. 
We compare a spectrum of first-stage retrieval methods, including lexical matching (BM25~\cite{robertson1994okapi} and TF-IDF~\cite{sparck1972statistical}), dense retrieval (BGE~\cite{xiao2024c} and Contriever~\cite{izacard2021unsupervised}), and a simple hybrid method~\cite{bruch2023analysis} that interleaves the ranked lists from BM25 and BGE to combine lexical and semantic signals. 
We further study LLM-based reranking~\cite{sun2023chatgpt} as a second-stage ranking mechanism: given a top-$50$ candidate pool retrieved by BM25, the target LLM reranks the candidates based on their relevance to the current request.

Tables~\ref{tab:retrieval_recall_merged} and ~\ref{tab:retrieval_N_merged} report Recall@$K$ and nDCG@$K$ across all benchmarks.
Several conclusions emerge clearly from the results. 
First, skill retrieval is already feasible, but far from solved. 
Across datasets, existing retrievers can achieve high Recall@10, confirming that skill descriptions do contain usable retrieval signals. 
However, retrieval quality varies sharply across benchmarks, and some settings remain difficult even for the strongest methods. 
This suggests that the difficulty of skill retrieval is highly task-dependent: in some domains, the relevant skill is expressed in ways that align naturally with skill text, while in others, the mapping from user request to useful skill is substantially more indirect.
Second, there is no universal winner among conventional retrievers. 
Sparse methods remain surprisingly competitive, especially when the required capability is exposed through distinctive terminology, formulas, or code-like surface patterns. 
Dense retrieval is stronger when relevance is expressed more semantically and less through direct lexical overlap, with BGE generally outperforming Contriever in this setting. 
Overall, these results suggest that skill retrieval is not well captured by either pure keyword matching or pure embedding similarity alone: the relevant signal is partly lexical, partly semantic, and often intertwined with procedural intent.
Third, combining lexical and semantic signals mainly improves candidate coverage rather than consistently identifying the best skill at the first rank. 
The hybrid retriever often strengthens top-$10$ recall, indicating that sparse and dense retrieval provide complementary candidates. 
Finally, LLM-based reranking is the strongest overall retrieval strategy. 
Once a candidate pool is available, an LLM can substantially improve the ranking quality. 
This suggests that skill retrieval depends not only on topical relevance, but also on recognizing whether a candidate actually constitutes an actionable capability for the current task. 
At the same time, reranking quality is not strictly monotonic with model scale, indicating that better skill retrieval depends on more than parameter count alone.

Overall, RQ3 yields a clear but qualified answer: existing retrievers can make skill retrieval feasible, but they are not reliable enough to make it a solved problem.
Sparse and dense methods exhibit complementary strengths, and LLM reranking provides the strongest overall performance. 
These findings establish retrieval as a meaningful bottleneck in SRA and motivate the next question: how does retrieval quality translate into end-to-end performance gains for SR-Agents?

\subsection{RQ4: How Does Retrieval Quality Affect End-Task Performance?}

\begin{table*}[t]
\caption{End-to-end performance (\%) under different retrievers across six SRA-Bench datasets. The best results are in bold, and the second-best results are underlined.}
\label{tab:end_to_end_retriever_comparison}
\vspace{1.5mm}
\centering
\small
\setlength{\tabcolsep}{5pt}
\renewcommand{\arraystretch}{1.15}
\resizebox{\textwidth}{!}{
\begin{tabular}{llccccccc}
\toprule
Model & Method & TheoremQA & LogicBench & ToolQA & CHAMP & MedCalc & BigCodeBench & Average \\
\midrule

\multirow{5}{*}{Llama-3.1-8B}
& BM25                     & 36.5 & \underline{58.3} & 13.6 & \underline{27.8} & 36.7 & 34.2 & 32.7 \\
& TF-IDF                   & 37.8 & 51.8 & 12.3 & \underline{27.8} & 40.0 & \underline{34.9} & 32.4 \\
& BGE                      & \underline{39.6} & 55.9 & 17.8 & 24.7 & \underline{42.3} & 34.1 & \underline{34.9} \\
& Contriever               & 37.8 & 54.7 & \textbf{19.7} & 24.2 & 37.5 & 31.3 & 33.4 \\
& BM25 + Rerank            & \textbf{40.8} & \textbf{60.4} & \underline{18.3} & \textbf{28.3} & \textbf{56.8} & \textbf{36.1} & \textbf{39.4} \\
\midrule

\multirow{5}{*}{Llama-3.3-70B}
& BM25                     & 60.1 & \underline{66.1} & 35.0 & \underline{61.9} & 59.5 & \textbf{52.9} & \underline{52.7} \\
& TF-IDF                   & 56.4 & 56.3 & 30.7 & 59.6 & 60.8 & \underline{51.4} & 49.6 \\
& BGE                      & \underline{61.7} & 58.6 & \underline{38.7} & \textbf{64.1} & \underline{60.9} & 49.9 & 52.6 \\
& Contriever               & 59.7 & 58.9 & 38.1 & 61.4 & 58.6 & 50.1 & 51.7 \\
& BM25 + Rerank            & \textbf{64.3} & \textbf{69.2} & \textbf{40.3} & 58.3 & \textbf{77.2} & \underline{51.4} & \textbf{58.3} \\
\midrule

\multirow{5}{*}{Mistral3.1-24B}
& BM25                     & 58.8 & \underline{57.2} & 34.4 & \textbf{59.6} & 54.6 & \underline{45.6} & \underline{48.5} \\
& TF-IDF                   & 57.7 & 45.7 & 29.9 & 56.1 & \underline{57.8} & 45.4 & 46.0 \\
& BGE                      & \underline{59.0} & 50.1 & \underline{36.3} & 55.6 & 56.9 & 45.3 & 48.3 \\
& Contriever               & 55.3 & 46.7 & 35.0 & \underline{57.0} & 48.4 & 43.0 & 44.8 \\
& BM25 + Rerank            & \textbf{62.0} & \textbf{64.1} & \textbf{38.7} & 52.0 & \textbf{74.6} & \textbf{48.4} & \textbf{55.4} \\
\midrule

\multirow{5}{*}{Qwen3-235B}
& BM25                     & \underline{64.4} & \underline{72.8} & 38.0 & 68.6 & 66.2 & \textbf{50.4} & 56.2 \\
& TF-IDF                   & 60.9 & 69.1 & 34.6 & 70.4 & 67.8 & \underline{50.1} & 54.6 \\
& BGE                      & 62.5 & 68.4 & 42.7 & \underline{72.2} & \underline{68.8} & 48.1 & \underline{56.7} \\
& Contriever               & 62.5 & 68.7 & \underline{42.9} & 71.7 & 65.5 & 50.0 & 56.5 \\
& BM25 + Rerank            & \textbf{66.3} & \textbf{80.1} & \textbf{45.2} & \textbf{74.0} & \textbf{82.2} & 49.4 & \textbf{62.6} \\
\midrule

\multirow{5}{*}{Qwen3-32B}
& BM25                     & \underline{68.1} & \underline{70.5} & 36.2 & \textbf{70.4} & 59.5 & \textbf{49.0} & 54.3 \\
& TF-IDF                   & 63.9 & 70.3 & 33.1 & 65.0 & 60.5 & 48.3 & 52.7 \\
& BGE                      & 67.1 & 70.0 & 40.0 & 65.5 & \underline{62.5} & 48.9 & \underline{55.5} \\
& Contriever               & 65.6 & 69.1 & \underline{40.6} & 57.8 & 58.6 & 48.7 & 54.1 \\
& BM25 + Rerank            & \textbf{69.3} & \textbf{78.3} & \textbf{43.0} & \underline{65.9} & \textbf{82.2} & \textbf{49.0} & \textbf{61.8} \\
\midrule

\multirow{5}{*}{Qwen3-4B}
& BM25                     & \underline{64.0} & \underline{69.1} & 30.2 & \textbf{68.6} & 36.1 & \textbf{41.5} & 45.5 \\
& TF-IDF                   & 59.7 & 65.7 & 24.3 & 47.1 & \underline{43.8} & 39.6 & 43.1 \\
& BGE                      & 62.5 & 66.1 & \textbf{38.6} & 61.0 & 43.1 & 38.9 & \underline{47.7} \\
& Contriever               & 57.7 & 65.1 & 36.2 & 59.2 & 38.9 & 39.4 & 45.4 \\
& BM25 + Rerank            & \textbf{65.5} & \textbf{72.9} & \underline{36.7} & \underline{65.0} & \textbf{66.2} & \underline{40.9} & \textbf{53.8} \\
\bottomrule
\end{tabular}
}
\end{table*}

To investigate how retrieval quality affects downstream task performance, we evaluate SR-Agents with five retrieval strategies: BM25~\cite{robertson1994okapi}, TF-IDF~\cite{sparck1972statistical}, BGE~\cite{xiao2024c}, Contriever~\cite{izacard2021unsupervised}, and BM25 + Rerank~\cite{sun2023chatgpt}. Given an input query, each method retrieves candidate skills, from which only the top-1 skill is selected and injected into the LLM for final answer generation. For BM25 + Rerank, we first retrieve the top 50 skills using BM25, then use the corresponding base LLM to rerank these candidates, and finally select the top-1 reranked skill for injection.

Table~\ref{tab:end_to_end_retriever_comparison} compares end-to-end SR-Agents' performance under different retrievers across six SRA-Bench datasets and six LLMs. Overall,  RQ4 does not admit a simple answer: stronger retrieval generally improves end-task performance, but the relationship is neither direct nor strictly monotonic. Across models and datasets, retrieval pipelines that achieve better ranking quality, especially higher precision at top ranks, tend to deliver stronger downstream performance, indicating that retrieval remains an important bottleneck for SR-Agents. In particular, reranking BM25-retrieved candidates often produces the most consistent end-to-end gains, suggesting that downstream success depends not only on whether relevant skills are retrieved, but also on whether they are reliably selected and utilized by the agent.

At the same time, the impact of retrieval quality varies substantially across tasks, and no single retrieval paradigm consistently dominates across benchmarks. This variability highlights a key distinction between skill retrieval and conventional document retrieval: in Skill Retrieval Augmentation, the retrieved units are executable capability packages rather than plain textual evidence, so end-task success depends not only on retrieval itself, but also on whether the agent can correctly interpret, incorporate, and execute the retrieved skills. As a result, improvements in standalone retrieval metrics do not always translate proportionally into gains on the final task. Better retrieval should therefore be understood as increasing the likelihood of successful skill use, rather than guaranteeing it.

Therefore, RQ4 suggests that retrieval quality is a necessary but insufficient condition for strong SRA performance. Improving retrieval is clearly beneficial, particularly when it enhances top-rank precision and reduces exposure to distracting or misleading skills. Still, its ultimate effect is mediated by downstream skill incorporation and execution. This further supports our broader claim that Skill Retrieval Augmentation should be studied not as a retrieval problem in isolation, but as a coupled pipeline in which retrieval and subsequent skill use jointly determine end-task outcomes.

\subsection{RQ5: Are Current SR-Agents Relevance-Aware in Skill Loading?}

\begin{table}[t]
\caption{Skill loading rate (\%) across models and datasets. Skill loading rate measures whether the agent successfully incorporates at least one valid skill before producing the final answer. 
Overall is computed over all 5,400 instances, rather than by averaging dataset-level rates.}
\label{tab:skill_loading_rate}
\vspace{1.5mm}
\centering
\small
\setlength{\tabcolsep}{5pt}
\renewcommand{\arraystretch}{1.1}
\begin{tabular}{lccccccc}
\toprule
Model & TheoremQA & LogicBench & ToolQA & CHAMP & MedCalc & BigCodeBench & Overall \\
\midrule
Qwen3-4B    & 15.5\% & 55.0\% & 2.4\%  & 13.5\% & 33.4\% & 14.8\% & 21.0\% \\
Llama-8B    & 97.9\% & 99.7\% & 2.9\%  & 86.5\% & 97.8\% & 96.0\% & 72.1\% \\
Mistral-24B & 45.1\% & 38.0\% & 29.6\% & 22.9\% & 32.1\% & 6.4\%  & 28.3\% \\
Qwen3-32B   & 24.9\% & 45.9\% & 0.6\%  & 1.8\%  & 41.3\% & 7.2\%  & 20.1\% \\
Llama-70B   & 31.7\% & 32.6\% & 6.9\%  & 1.8\%  & 0.9\%  & 0.0\%  & 11.1\% \\
Qwen3-235B  & 77.2\% & 87.5\% & 1.5\%  & 15.7\% & 56.4\% & 88.2\% & 54.1\% \\
GLM-5.1     & 55.3\% & 25.7\% & 7.8\%  & 25.6\% & 86.5\% & 27.5\% & 37.8\% \\
GPT-5.4     & 6.8\%  & 0.4\%  & 38.7\% & 15.7\% & 74.7\% & 19.5\% & 31.2\% \\
\bottomrule
\end{tabular}

\end{table}

\begin{table}[t]
\caption{Comparison of overall skill-loading rates between instances whose gold skill is covered by BM25 top-50 and those whose gold skill is not. Instances are grouped by whether at least one gold skill appears in BM25 top-50 for multi-label datasets.}
\label{tab:gold_availability_loading}
\vspace{1.5mm}
\centering
\small
\setlength{\tabcolsep}{6pt}
\renewcommand{\arraystretch}{1.1}
\begin{tabular}{lccc}
\toprule
Model & Gold in Top-50 Load Rate & Gold not in Top-50 Load Rate & Diff. \\
\midrule
Qwen3-4B    & 21.8\% & 16.9\% & +4.9pp  \\
Llama-8B    & 73.7\% & 64.1\% & +9.6pp  \\
Mistral-24B & 30.4\% & 17.3\% & +13.1pp \\
Qwen3-32B   & 21.2\% & 14.3\% & +6.9pp  \\
Llama-70B   & 10.4\% & 14.5\% & -4.2pp  \\
Qwen3-235B  & 57.8\% & 35.3\% & +22.4pp \\
GLM-5.1     & 43.5\% & 8.2\%  & \textbf{+35.4pp }\\
GPT-5.4     & 36.6\% & 3.6\%  & \underline{+33.0pp} \\
\bottomrule
\end{tabular}

\end{table}

Table~\ref{tab:skill_loading_rate} and Table~\ref{tab:gold_availability_loading} examine whether current LLMs load skills in a relevance-aware manner, namely, whether they are more likely to load a skill when the retrieved candidate set actually contains a gold skill.
Overall, the answer is largely negative. Skill-loading behavior is highly model-dependent rather than a stable capability that improves monotonically with scale. Under the same skill corpus and retriever, different models exhibit sharply different loading propensities across datasets: some tend to load skills almost indiscriminately, while others are overly conservative and rarely load at all. This already suggests that current SR-Agents do not follow a consistent or rational loading policy, and that skill incorporation is heavily shaped by model-specific tendencies.

More importantly, for most open-source models, the presence of a gold skill in the retrieved candidate pool has only limited influence on whether the agent decides to load a skill. Although gold-covered cases sometimes lead to slightly higher loading rates, many models still load skills at substantial rates even when no gold skill is present in the BM25 top-50. This reveals a clear form of \emph{skill-loading hallucination}: rather than first determining whether the retrieved candidates contain a genuinely relevant capability, the model often proceeds to load some retrieved skill anyway. As a result, successful gold-skill loading is often accompanied by a comparable tendency to load irrelevant skills, exposing a substantial disconnect between retrieval success and effective utilization. A clearer form of relevance awareness emerges only in the strongest models. Among the open-source models, this pattern becomes apparent only at the 235B scale. By contrast, frontier models such as GLM-5.1 and GPT-5.4 show much sharper separation between gold-covered and gold-absent cases, indicating a substantially stronger ability to condition skill loading on actual candidate relevance. Importantly, their advantage does not come from indiscriminately loading more skills overall, but from being more selective: they are more willing to load when a relevant skill is available, and much less willing to do so when it is not.

Taken together, these results show that most current LLMs are generally not reliably relevance-aware when incorporating skills. 
Although frontier models such as GLM-5.1 and GPT-5.4 exhibit stronger separation between gold-covered and gold-absent cases than open-source models, their behavior remains far from ideal: skill loading is still not fully governed by whether a genuinely relevant skill has been retrieved. 
For most models, whether a suitable skill has actually been retrieved exerts surprisingly weak control over loading behavior, leaving a major gap between successful retrieval and effective skill utilization.

\subsection{RQ6: Are Current SR-Agents Need-Aware in Skill Loading?}

\begin{table*}[t]
\caption{Need-aware skill-loading analysis. We partition instances according to whether the same model can solve them \emph{without} external skills (\textit{Skill-Free Correct}) or not (\textit{Skill-Free Wrong}). A need-aware SR-Agent should load skills more often when the task cannot be solved natively. $\Delta$ is computed as \textit{Skill-Free Wrong} $-$ \textit{Skill-Free Correct}, in percentage points (pp).}
\label{tab:need_aware_loading}
\vspace{1mm}
\centering
\small
\setlength{\tabcolsep}{5pt}
\renewcommand{\arraystretch}{1.15}
\resizebox{\textwidth}{!}{

\begin{tabular}{lcccccccc}
\toprule
& \multicolumn{3}{c}{Skill-Free Correct} & \multicolumn{3}{c}{Skill-Free Wrong} & \multicolumn{2}{c}{Difference} \\
\cmidrule(lr){2-4} \cmidrule(lr){5-7} \cmidrule(lr){8-9}
Model & $n$ & Load Rate & Gold Load Rate& $n$ & Load Rate & Gold Load Rate& $\Delta$ Load & $\Delta$ Gold \\
\midrule
Qwen3-4B   & 1618 & 23.6 & 13.1 & 2902 & 20.8 & 17.2 & $-2.8$  & $+4.1$ \\
Llama-8B   & 1273 & 84.5 & 59.7 & 3247 & 69.4 & 54.0 & $-15.1$ & $-5.7$ \\
Mistral-24B& 1893 & 28.0 & 22.0 & 2627 & 32.1 & 25.0 & $+4.1$  & $+3.0$ \\
Qwen3-32B  & 2201 & 21.4 & 17.5 & 2319 & 21.0 & 18.8 & $-0.4$  & $+1.3$ \\
Llama-70B  & 2105 & 10.7 & 7.6  & 2415 & 10.1 & 7.1  & $-0.6$  & $-0.6$ \\
Qwen3-235B & 2335 & 59.5 & 46.3 & 2185 & 56.0 & 46.6 & $-3.5$  & $+0.4$ \\
GLM-5.1    & 2616 & 45.0 & 42.5 & 1904 & 41.5 & 38.0 & $-3.4$  & $-4.4$ \\
GPT-5.4    & 2790 & 34.4 & 33.2 & 1730 & 40.2 & 37.0 & $+5.8$  & $+3.8$ \\
\midrule
Overall    & 16831 & 36.9 & 30.0 & 19329 & 36.9 & 30.5 & $+0.1$  & $+0.5$ \\
\bottomrule
\end{tabular}
}
\end{table*}

We next examine whether current LLMs exhibit \emph{need-aware} skill-loading behavior during incorporation. 
Intuitively, a rational SR-Agent should be more inclined to load external skills when a task exceeds its native parametric capability, and should remain more conservative when the same task can already be solved without external skill augmentation. 
To test this, we partition instances based on whether the same model answers them correctly in the skill-free setting, and then compare skill-loading behavior between natively solvable and natively unsolvable cases. 
To isolate the effect of need awareness from retrieval availability, this analysis is restricted to instances where a gold skill appears among the top-50 retrieved candidates, ensuring that both groups are evaluated under the same condition in which relevant external skills are available for loading. 
The resulting loading rates are shown in Table~\ref{tab:need_aware_loading}.

The results reveal a striking absence of need awareness. Across models, skill-loading rates remain remarkably similar between instances that the model can already solve on its own and those it fails to solve without external skills, including the frontier models that showed comparatively stronger relevance awareness in RQ5. In other words, current LLMs do not meaningfully condition skill-loading decisions on whether a genuine capability gap is present. Even when a task demonstrably exceeds the model's native competence, the agent is generally no more likely to load a skill than when the task is already within reach of its parametric knowledge. This pattern indicates that skill loading is not functioning as a targeted compensatory mechanism for missing capability, but is instead triggered in a largely indiscriminate manner.

Taken together, these results show that current LLMs are generally not need-aware when incorporating skills. Combined with the weak relevance awareness observed in RQ5, this points to a deeper limitation of today's SR-Agents: they still lack a reliable mechanism for deciding what external skill to load and when external loading is actually needed. This substantially weakens the practical value of large-scale skill retrieval augmentation and suggests that scalable skill augmentation is fundamentally a problem of controlled, selective, and need-aware skill utilization, rather than retrieval alone.

\section{Related Work} 

\subsection{Retrieval Augmented Generation and Tool Use} 

Retrieval-Augmented Generation (RAG) extends large language models (LLMs) by grounding them with task-relevant information retrieved from external knowledge sources~\cite{lewis2020retrieval,dong2025decoupling,tu2025robust,su2025dynamic,tu2026analytical}. 
A growing body of work has shown that RAG is effective for reducing hallucinations~\cite{wang2026joint,su2025towards,su2024unsupervised,su2025surge}, supporting knowledge updating~\cite{wang2025decoupling,wang2025knowledge}, and adapting LLMs to new domains without costly full-model retraining~\cite{su2025judge,su2024stard}. 
Most existing RAG systems follow a retrieval-then-read pipeline, in which the input query is first used to retrieve relevant documents from a large external corpus~\cite{robertson2009probabilistic,su2024wikiformer,fang2024scaling,su2025pre}, and the retrieved evidence is then incorporated into the model's context to assist with generation. 
Beyond this standard formulation, recent extensions have explored dynamic RAG~\cite{jiang2023active,su2024dragin,su2024mitigating}, graph RAG~\cite{edge2024local}, parametric RAG~\cite{su2025parametric,tan2025dynamic}, and agentic RAG~\cite{jin2025search}.

In parallel, a rich line of work explores tool use and function calling, in which models or agents learn to invoke external APIs or functions. These studies typically focus on tool/function use rather than the retrieval of reusable skills~\cite{tang2026multi,schick2023toolformer}. This paradigm is well exemplified by models such as Toolformer \cite{schick2023toolformer}, Gorilla \cite{patil2024gorilla}, and ToolLLM \cite{qin2023toolllm}, and is evaluated on benchmarks such as BFCL \cite{patil2025bfcl}.

While these paradigms have dramatically expanded model capabilities, SRA operates at their intersection while systematically addressing their respective limitations. 
Unlike RAG, which retrieves static knowledge, SRA retrieves executable and operational capabilities. Unlike conventional tool use, which assumes a visible, limited set of APIs, SRA retrieves from an open-ended corpus of modular packages that encompass instructions, invocation conditions, and procedural guidance. Consequently, SRA introduces a more complex pipeline that requires evaluating retrieval relevance alongside downstream skill incorporation and task application. 

\subsection{Agent Skills} 
Beyond atomic tools, recent research increasingly treats modular skills as a core abstraction for agent design. Pioneering systems like Voyager \cite{wang2023voyager} demonstrate the value of maintaining an ever-growing skill corpus of executable programs for embodied lifelong learning. 
Furthermore, recent analyses \cite{ling2026agentskills} have begun to map the emerging ecosystem of agent skills as plug-and-play extensions for LLMs. 
While these works underscore the utility of reusable skill abstractions, their primary focus remains on skill acquisition, code synthesis, or ecosystem analysis. SRA departs from this focus by addressing the critical bottleneck of scalability: rather than investigating how to create or store skills, we tackle the challenge of retrieval-time access, investigating how agents can dynamically identify and load the right capabilities from a massive, out-of-context library. 

\subsection{Benchmarking Skill Effectiveness} 
As skill ecosystems continue to expand, rigorous benchmarks are increasingly needed to evaluate how effectively external skills support agent performance. 
However, existing benchmarks primarily evaluate tool/function invocation, and even when retrieval is involved, they do not decompose the full pipeline of standalone skill retrieval, skill incorporation, and downstream task application. 
More recently, SkillsBench \cite{li2026skillsbench} was introduced to study whether curated skills genuinely assist agents in solving diverse tasks. While highly complementary, our work addresses a distinct and fundamentally different problem. By formulating \emph{skill retrieval augmentation} as a standalone research problem, we introduce SRA-Bench, which provides a fine-grained, decomposed evaluation pipeline. Instead of merely assessing whether skills are broadly beneficial, SRA-Bench diagnoses the specific bottlenecks of scalable augmentation: whether the correct skills can be retrieved from a vast corpus, whether the agent can selectively incorporate them without being overwhelmed, and whether this multi-stage process ultimately translates into measurable downstream success.

\section{Toward a Research Agenda for Skill Retrieval Augmentation}

The results of this paper suggest that Skill Retrieval Augmentation (SRA) should not be understood merely as a new retrieval problem over skills, but as a broader research agenda for scalable capability augmentation in agent systems.
On the positive side, our experiments show that retrieving external skills from a large skill corpus can substantially improve downstream performance, confirming the practical promise of augmenting agents with large-scale reusable capabilities beyond their native parametric knowledge.
At the same time, our findings also make clear that retrieval alone is far from sufficient.
Even when relevant skills are retrieved, current agents often struggle to identify which skills are actually useful, decide whether external help is needed, and reliably convert retrieved skills into better task execution.
Taken together, these observations suggest that the next stage of progress in SRA will require advances not only in retrieval methods but also in how skills are organized, maintained, incorporated, internalized, and continually improved over time.
In this section, we outline several research directions that, in our view, define the emerging agenda of SRA.

\paragraph{From unstructured skill collections to structured skill libraries.}
The first research direction concerns how a large-scale skill corpus should be represented and organized.
In this paper, we instantiate SRA with a large, noisy skill collection, enabling us to study retrieval and skill use under realistic open-world conditions.
However, simply treating a collection of skills as an unstructured list of independently indexed skills may become increasingly inadequate as the number of skills grows.
Unlike ordinary documents, skills are reusable capability units with rich relationships among themselves.
Some skills serve as prerequisites for others, some provide alternative implementations of the same capability, some specialize broader skills into narrower domains, and some are naturally composed into multi-step workflows.
As a result, future SRA systems may need more structured forms of organization, such as graphs, hierarchies, clusters, or explicit dependency structures over skills.
These structures could help retrieval systems narrow the search space, reduce interference from superficially similar yet functionally irrelevant skills, and support more effective reasoning across the available capabilities.
Importantly, our proposed large-scale skill corpus and gold-skill annotations introduced in this paper provide a concrete testbed for studying these questions under decomposed evaluation.

\paragraph{Skill quality control, offline refinement, and skill evolution.}
The second direction concerns the quality of the skill corpus itself. Open skill ecosystems are inherently heterogeneous: many skills may be incomplete, poorly written, outdated, redundant, or even incorrect. This challenge is qualitatively different from conventional document retrieval, because the retrieved unit in SRA is not merely textual evidence but an actionable capability package whose errors may directly mislead downstream reasoning and execution. As a result, scalable SRA will likely require dedicated pipelines for skill validation, debugging, and refinement. One promising direction is offline skill optimization: given abundant offline compute, a system may iteratively inspect skills, detect missing preconditions, repair procedural flaws, rewrite ambiguous descriptions, attach executable resources, or revise invocation conditions to make the skill more retrievable and more usable. Such optimization could be driven by static analysis of the skill artifact itself, by execution-based testing on synthesized tasks, or by agentic self-improvement loops in which agents attempt to use a skill, reflect on failure cases, and revise the skill accordingly. In this view, the skill corpus is no longer a passive repository, but a continually maintained capability resource whose quality can be improved over time.

\paragraph{From semantic matching to utility-aware skill retrieval.}
The third direction concerns retrieval itself. Our formulation already emphasizes that retrieval in SRA should be evaluated not only by semantic relevance, but by downstream utility: whether the retrieved candidates actually help solve the task. 
This suggests that future skill retrieval methods should be optimized for expected usefulness rather than topical similarity alone. 
In contrast to classical document retrieval, where the target is often factual evidence, skill retrieval requires identifying actionable capability packages whose value depends on whether they can be effectively incorporated and operationalized by the agent. 
This opens up several opportunities. 
Retrieval models could be trained directly from end-task feedback, using success or failure on downstream tasks as a supervision signal. Dense retrievers could be optimized with utility-aware objectives rather than purely relevance-based contrastive signals. 
Rerankers could prioritize skills that are not only semantically related to the query but also likely to close the agent's current capability gap. 
More ambitiously, retrieval may need to move beyond single-skill ranking to retrieve sets of complementary skills, or to be conditioned on the agent's intermediate reasoning state rather than only on the initial user request. 
Taken together, these challenges suggest that skill retrieval in SRA should be studied as a distinct IR problem, rather than as a straightforward extension of classical document retrieval.

\paragraph{Toward Parametric Skill Augmentation.}
Finally, a particularly promising direction is {Parametric Skill Augmentation}, where external skills are not only retrieved as text, but also transformed offline into plug-in parameters that can be loaded into the model when needed. This direction is motivated by recent advances in Parametric RAG~\cite{su2025parametric,su2025sigir}, which suggest that external information need not be injected only through the input context, but can instead be integrated into the model's parameters, thereby reducing online context overhead while enabling deeper interaction with the model's internal computation. 
For skills, this idea may be even more natural than for documents. Unlike open-ended world knowledge, skills are reusable capability modules: many are invoked repeatedly across tasks, and their role is not merely to provide declarative information, but to induce reliable procedures, decision patterns, and action policies. 
As a result, repeatedly exposing the same skill through textual descriptions can be an inefficient use of context budget, especially for frequently used skills. Parameterizing such head skills could amortize this cost by replacing repeated token-level injection with compact plug-in modules, while retaining rare, long-tail, or rapidly evolving skills in an external retrieval corpus. 
Beyond efficiency, parameterization may also improve effectiveness. As these plug-in parameters are constructed offline, they can be explicitly optimized for downstream use, for example, through task-aware initialization, refinement, or feedback-driven updates, rather than being presented to the model only as surface-form prompt text. 
More importantly, in-context skill injection primarily affects the model at the input level. In contrast, parametric skill augmentation offers a potential pathway for agents to internalize external capabilities more deeply into the LLM's parameters that drive reasoning and acting. In this sense, parametric skill augmentation is not merely a compression strategy for SRA but a potentially stronger mechanism for turning retrieved skills into durable, operational competence. A hybrid architecture may therefore be especially attractive: frequently used core skills could be maintained as parametric modules for efficient and reliable access, while the open-ended long tail of skills remains retrievable from an external corpus.

% Taken together, these directions suggest that SRA is not a narrowly scoped extension of retrieval-augmented generation, but an emerging research paradigm for scalable capability augmentation, for which the resources and findings introduced in this paper provide a concrete starting point.

\section{Conclusion}

In this paper, we formulate {Skill Retrieval Augmentation (SRA)} as a new paradigm for augmenting agents with external capabilities at scale. 
Rather than assuming that all potentially useful skills can be explicitly exposed in context, SRA studies how agents should retrieve, incorporate, and apply reusable skills from large external skill corpora on demand. 
To make this problem concrete and measurable, we construct a large-scale {skill corpus}, introduce {SRA-Bench} as the first benchmark for decomposed evaluation of the SRA pipeline, and establish {SR-Agents} as a baseline family for systematic empirical study. 
Together, these contributions provide a concrete foundation for studying scalable skill augmentation as a distinct problem in agent systems.

Our experiments show both the promise and the central challenge of this paradigm. 
On the one hand, external skills can substantially improve downstream performance when they are correctly accessed and used. 
On the other hand, our results reveal that scalable skill augmentation cannot be solved by retrieval alone. 
Even when relevant skills are successfully retrieved, current LLM agents often still fail to recognize whether external help is needed, which retrieved skills are truly worth loading, and how to translate those skills into reliable gains in reasoning and action. 
These findings suggest that the core bottleneck of SRA lies not only in finding relevant capabilities but in enabling agents to access them selectively, incorporate them appropriately, and operationalize them robustly.

We therefore view this work not merely as introducing a benchmark, but as helping define a broader research agenda. 
By releasing the skill corpus, SRA-Bench, and the analyses in this work, we aim to provide a shared foundation for future research on how external skills should be organized, maintained, retrieved, selected, refined, internalized, and accumulated over time. 
More broadly, if Retrieval-Augmented Generation made \emph{external knowledge} a central object of study for language models, we believe Skill Retrieval Augmentation can help make \emph{external capabilities} a central object of study for agent systems. 
The next generation of agents will be shaped not only by what they know but also by how effectively they can access and leverage capabilities beyond their weights.

\bibliographystyle{plain}
\bibliography{ref}

\clearpage

\appendix

\section{Dataset-Specific Construction Details}
\label{sec:appendix-construction}

This appendix provides the dataset-specific construction details underlying SRA-Bench.
While Section~\ref{sec:golden_skill_construction} presents the benchmark construction pipeline at a unified level, the actual supervision signals differ markedly across source datasets.
For example, reusable capabilities may be anchored by theorem names, logic patterns, tool workflows, medical calculator types, mathematical concepts, or software libraries.
To make the construction process fully transparent, we therefore unpack the pipeline separately for each dataset.
For every source benchmark, we describe the full path from raw source-side annotations to finalized gold skills.
This includes the available source materials, the dataset-specific inputs used to construct an initial LLM draft, the expert revision process that converts the draft into a reusable, leakage-controlled skill artifact, and concrete construction examples that illustrate the annotation results.
These details are important for understanding how heterogeneous instance-level signals are abstracted into a unified form of gold-skill that is suitable for retrieval, incorporation, and end-to-end evaluation in the SRA setting.

Beyond documentation, this appendix also serves as evidence for a central design choice in SRA-Bench:
gold skills are not directly inherited from the source datasets, but manually constructed from structured supervision signals into standalone capability artifacts.
By presenting the construction procedure at the per-dataset level, we aim to make this process easier to inspect, verify, and reproduce.

% Appendix: TheoremQA Construction Details
\subsection{TheoremQA}
\label{sec:appendix-theoremqa}

% --- A.x.1 Source Material ---
% \subsubsection{Source Material}
% \label{sec:appendix-theoremqa-source}

Each instance in TheoremQA contains a \emph{question}, a \emph{theorem annotation} identifying the relevant theorem, and evaluation metadata including the ground-truth \emph{answer} and its \emph{answer type} (float, integer, bool, list, or option).
The original dataset contains approximately 800 instances spanning 334 theorems.
We remove image-dependent questions and retain the text-only subset, yielding 747 instances associated with 320 theorems.

The retained theorems span four STEM domains: mathematics (57\%), EECS (17\%), physics (15\%), and finance (11\%).
Each theorem is further tagged with a \emph{field} and \emph{subfield} (e.g., Mathematics / Combinatorics), which provides useful annotation-level context during skill construction.
Each instance is associated with exactly one theorem, and each theorem is mapped to one gold skill.

\subsubsection{LLM Draft Generation}
\label{sec:appendix-theoremqa-draft}

For each theorem, the LLM receives the theorem name, its field and subfield, all associated questions and answers, and domain-appropriate reference pages (e.g., Wikipedia and MathWorld\footnote{\url{https://mathworld.wolfram.com}, a comprehensive mathematics reference maintained by Wolfram Research.} for mathematics topics) as input.
Since the dataset does not provide solution steps, the draft must infer a preliminary application procedure from the theorem label, the distribution of associated questions, and external reference materials. The questions and answers are used only to help the LLM characterize the scope of application during drafting; any answer-revealing content, benchmark-specific shortcuts, or copied examples are removed during expert revision under the leakage-control principle described in Section~\ref{sec:golden_skill_construction}.
The following prompt template is used:

\begin{promptbox}[Prompt Template for TheoremQA]
\small\ttfamily
You are a STEM education expert. Given a theorem and a set of problems that apply it, write a reusable skill document that teaches how to apply this theorem to solve new problems.

\medskip
\textbf{Theorem:} \{theorem\_name\} \\
\textbf{Field:} \{field\} \quad \textbf{Subfield:} \{subfield\}

\textbf{Reference material:} \\
\{reference\_pages\}

\textbf{Problems applying this theorem} (each with its ground truth answer): \\
\{questions\_and\_answers\}

\medskip
Based on the above materials, write a skill document with the following structure:
\begin{enumerate}[leftmargin=*]
    \item \textbf{name}: Canonical theorem name.
    \item \textbf{description}: One sentence summarizing what the theorem computes or states.
    \item \textbf{Content} (Markdown):
    \begin{itemize}
        \item Core principle or theorem statement.
        \item When to recognize that this theorem applies.
        \item Step-by-step application procedure.
        \item Common errors or pitfalls.
        \item One worked example using a new problem (not from the problems above).
    \end{itemize}
\end{enumerate}
\end{promptbox}

\subsubsection{Expert Revision}
\label{sec:appendix-theoremqa-revision}

Beyond the general revision principles described in Section~\ref{sec:golden_skill_construction}, TheoremQA requires special attention to two recurring failure modes.

\begin{itemize}[leftmargin=*]
    \item \textbf{Procedural under-specification.}
    As the source data do not provide solution steps, LLM drafts often explain \emph{what} a theorem states but not \emph{how} to apply it in problem solving.
    As a result, many drafts resemble encyclopedia-style summaries, listing formulas and variable definitions without a clear procedure for application.
    During expert revision, such drafts are restructured into step-by-step skills, with explicit identification of inputs, case distinctions, and computation templates where needed.

    \item \textbf{Confusion with closely related concepts.}
    Many STEM theorems have neighboring variants or closely related formulas that are easy to confuse in practice (e.g., Lah numbers vs.\ Stirling numbers, population vs.\ sample covariance).
    LLM drafts may mention such related concepts without clearly distinguishing the conditions under which they apply.
    Expert revision, therefore, adds explicit disambiguation cues, comparison rules, and recognition patterns to reduce misapplication.
\end{itemize}

\subsubsection{Construction Examples}
\label{sec:appendix-theoremqa-examples}

We present two complete construction examples from different STEM domains.

% -------------------------------------------------------------------
\paragraph{Example 1: Dividend Discount Model (DDM).}
DDM has three common variants (zero growth, constant growth, and multi-stage); correctly selecting and applying the right variant is a challenge.

\begin{rawbox}[Source Material]
\textbf{Theorem:} Dividend Discount Model (DDM) \\
\textbf{Field:} Finance \quad \textbf{Subfield:} Asset pricing \\
2 of 5 associated instances:

\medskip
\textbf{Q:} If a stock pays a \$5 dividend this year, and the dividend has been growing 6\% annually, what will be the stock's intrinsic value, assuming a required rate of return of 12\%? \\
\textbf{A:} 88.33 \quad (float)

\textbf{Q:} CheckMate forecasts that its dividend will grow at 20\% per year for the next four years before settling down at a constant 8\% forever. Dividend (current year, 2016) = \$12; expected rate of return = 15\%. What is the fair value of the stock now? \\
\textbf{A:} 273.0 \quad (float)
\end{rawbox}

\begin{draftbox}[LLM Drafted Skill]
\texttt{name:} Dividend Discount Model (DDM) \\
\texttt{description:} Value stocks using dividend discount models including the Gordon Growth Model. \\[2pt]
\tcbline

\textbf{Dividend Discount Model (DDM)}

The DDM values a stock as the present value of all future dividends: $P_0 = \sum D_t / (1{+}r)^t$.

\textbf{Zero Growth (Preferred Stock):} $P_0 = D / r$.

\textbf{Constant Growth (Gordon Growth Model):}
$$P_0 = \frac{D_1}{r - g} = \frac{D_0 \times (1+g)}{r - g}$$
where $D_1$ is next period's dividend. Requires $r > g$.

\textbf{Multi-Stage Growth:}
For high initial growth $g_1$ for $n$ years transitioning to stable growth $g_2$, compute dividends during the high-growth phase and add the terminal value $\text{TV}_n = D_{n+1} / (r - g_2)$.

\hfill \textcolor{red}{$\triangleright$ \textit{No step-by-step procedure for multi-stage computation}}

\textbf{Worked Example:}
A stock pays \$5, growing 6\% annually, required return 12\%.
$P_0 = 5 \times 1.06 \;/\; (0.12 - 0.06) = 88.33$. \hfill \textcolor{red}{$\triangleright$ \textit{Same parameters as an input instance}}

[\ldots]
\end{draftbox}

\begin{skillbox}[Golden Skill]
\texttt{name:} Dividend Discount Model (DDM) \\
\texttt{description:} Valuing stocks using constant-growth and multi-stage dividend discount models, including terminal value computation. \\[2pt]
\tcbline

\textbf{Dividend Discount Model (DDM)}

The intrinsic value of a stock equals the present value of all future dividends: $P_0 = \sum D_t / (1{+}r)^t$. The formula simplifies according to the dividend growth pattern.

\textbf{Three Common Cases:}
\begin{itemize}[leftmargin=*]
    \item \textbf{Zero growth} (preferred stock): $P_0 = D / r$.
    \item \textbf{Constant growth} (Gordon): $P_0 = D_1 / (r {-} g)$, where $D_1 = D_0(1{+}g)$. \\
    Common error: using $D_0$ instead of $D_1$ in the numerator.
    \item \textbf{Multi-stage}: high growth $g_1$ for $n$ years, then stable growth $g_2$ forever.
\end{itemize}

\textbf{Multi-Stage Step-by-Step Procedure:}
\begin{enumerate}[leftmargin=*]
    \item Compute each high-growth dividend: $D_t = D_0 (1{+}g_1)^t$ for $t = 1, \ldots, n$.
    \item Discount each to present: $\text{PV}_t = D_t / (1{+}r)^t$.
    \item Terminal value at year $n$: $\text{TV}_n = D_n(1{+}g_2) / (r {-} g_2)$. \\
    \emph{Note:} the numerator is $D_{n+1} = D_n(1{+}g_2)$, \emph{not} $D_n$.
    \item Discount terminal value: $\text{PV(TV)} = \text{TV}_n / (1{+}r)^n$.
    \item Sum: $P_0 = \sum \text{PV}_t + \text{PV(TV)}$.
\end{enumerate}

\textbf{Critical Pitfalls:}
\begin{itemize}[leftmargin=*]
    \item Wrong terminal value base: TV uses $D_{n+1}$, not $D_n$.
    \item Missing growth step: given $D_0$, the first dividend is $D_1 = D_0(1{+}g)$, not $D_0$.
    \item Rate format: if $r$ and $g$ are given as percentages (e.g., 12\%), convert to decimals (0.12) before computing $r - g$.
\end{itemize}

\textbf{Worked Example:}
$D_0 = 4$, $g_1 = 18\%$ for 3 years, then $g_2 = 6\%$, $r = 14\%$.
$D_1 = 4.72$, $D_2 = 5.57$, $D_3 = 6.57$; PV of dividends = 12.86.
$D_4 = 6.57 \times 1.06 = 6.96$; $\text{TV}_3 = 6.96 / 0.08 = 87.0$; $\text{PV(TV)} = 58.7$.
$P_0 = 12.86 + 58.7 \approx \mathbf{71.6}$.
\end{skillbox}

\noindent\textbf{Revision summary.}
(1)~The LLM draft correctly presents the DDM formulas for all three cases, but treats multi-stage growth in a single sentence without a structured computation procedure.
The expert restructured the skill with an explicit five-step multi-stage template and documented the critical $D_{n+1}$ vs.\ $D_n$ pitfall by consulting the Wikipedia DDM reference.
(2)~The draft's example uses parameters identical to an input instance.
The expert replaced it with a newly constructed multi-stage example.

% -------------------------------------------------------------------
\paragraph{Example 2: Lah Numbers.}
Lah numbers count partitions of a set into non-empty ordered lists, and are frequently confused with the closely related Stirling numbers of the second kind.

\begin{rawbox}[Source Material]
\textbf{Theorem:} Lah Numbers \\
\textbf{Field:} Mathematics \quad \textbf{Subfield:} Combinatorics \\
2 of 6 associated instances:

\medskip
\textbf{Q:} How many ways are there to divide a set of 8 elements into 5 non-empty ordered subsets? \\
\textbf{A:} 11760 \quad (integer)

\textbf{Q:} A group of 7 people is to be divided into 3 committees. Within each committee, people are ranked in a certain order. In how many ways can this be done? \\
\textbf{A:} 12600 \quad (integer)
\end{rawbox}

\begin{draftbox}[LLM Drafted Skill]
\texttt{name:} Lah Numbers \\
\texttt{description:} Compute Lah numbers for counting partitions of a set into non-empty ordered subsets. \\
% \tcbline

\textbf{Lah Numbers}

The Lah number $L(n, k)$ counts the number of ways to partition a set of $n$ elements into $k$ non-empty ordered subsets (sequences).

\textbf{Formula:} $L(n, k) = \frac{n!}{k!} \binom{n-1}{k-1}$

\textbf{Procedure:}
\begin{enumerate}[leftmargin=*]
    \item Identify $n$ (total elements) and $k$ (number of groups).
    \item Compute $L(n,k) = \frac{n!}{k!} \times \binom{n-1}{k-1}$.
\end{enumerate}

\textbf{Special Cases:} $L(n,1) = n!$; \; $L(n,n) = 1$.

\hfill \textcolor{red}{$\triangleright$ \textit{No distinction from Stirling numbers; models confuse ordered vs.\ unordered partitions}}

\textbf{Worked Example:}
Divide 8 elements into 5 ordered subsets:
$L(8,5) = \frac{8!}{5!} \times \binom{7}{4} = 336 \times 35 = 11760$. \hfill \textcolor{red}{$\triangleright$ \textit{Same parameters as an input instance}}
\end{draftbox}

\begin{skillbox}[Golden Skill]
\texttt{name:} Lah Numbers \\
\texttt{description:} Computing Lah numbers for counting ordered partitions of a set into non-empty ordered sequences. \\[2pt]
\tcbline

\textbf{Lah Numbers}

The Lah number $L(n, k)$ counts the number of ways to partition a set of $n$ elements into $k$ non-empty \textbf{ordered} subsets (sequences), where the order within each subset matters.

\textbf{Formula:}
$L(n, k) = \frac{n!}{k!} \binom{n-1}{k-1}$

\textbf{Step-by-Step Procedure:}
\begin{enumerate}[leftmargin=*]
    \item Identify $n$ (total elements) and $k$ (number of groups).
    \item Verify the problem asks for \textbf{ordered} groups (sequences, ranked committees, lists)---not unordered subsets.
    \item Compute $L(n,k) = \frac{n!}{k!} \times \binom{n-1}{k-1}$.
    \item Sanity check: $L(n,1) = n!$, \; $L(n,n) = 1$.
\end{enumerate}

\textbf{Distinguishing from Related Problems:}

\medskip
\begin{tabular}{lcc}
\toprule
\textbf{Concept} & \textbf{Groups ordered?} & \textbf{Within-group order?} \\
\midrule
Lah number $L(n,k)$ & No & Yes \\
Stirling 2nd kind $S(n,k)$ & No & No \\
$k! \times S(n,k)$ & Yes & No \\
\bottomrule
\end{tabular}

\medskip
\noindent\textbf{Key distinction}: ``subsets'' or ``groups'' (unordered within) $\to$ Stirling numbers.
``Sequences,'' ``ranked lists,'' or ``ordered subsets'' $\to$ Lah numbers.

\textbf{Worked Example:}
Divide 4 books into 2 non-empty ordered shelves:
$L(4, 2) = \frac{4!}{2!} \times \binom{3}{1} = 12 \times 3 = \mathbf{36}$.
\end{skillbox}

\noindent\textbf{Revision summary.}
(1)~The LLM draft provides the correct formula but does not distinguish Lah numbers from the closely related Stirling numbers of the second kind, which count \emph{unordered} partitions.
The expert added the comparison table and recognition patterns, which are critical since problems with nearly identical phrasing (e.g., ``groups'' vs.\ ``ordered groups'') require different formulas.
(2)~The draft's worked example uses parameters identical to an input instance ($n{=}8$, $k{=}5$).
The expert replaced it with a smaller, independently constructed example ($n{=}4$, $k{=}2$) that is easier to verify by hand.

\subsection{LogicBench}
\label{sec:appendix-logicbench}

Each LogicBench instance consists of a natural-language \emph{question} that poses a logical reasoning problem and a ground-truth \emph{answer}: either \texttt{yes}/\texttt{no} for binary question answering (BQA, 71\%) or \texttt{choice\_N} for four-option multiple-choice question answering (MCQA, 29\%).
The original benchmark contains 25 logic patterns across three categories: 9 propositional, 8 first-order, and 8 non-monotonic.
Among them, six first-order patterns are direct counterparts of propositional ones (e.g., first-order modus tollens tests the same underlying inference rule as propositional modus tollens).
We therefore exclude these redundant variants and retain 19 distinct patterns, yielding 760 instances in total.
Each retained pattern corresponds to a named inference rule, such as modus tollens, disjunctive syllogism, or default reasoning with exceptions, and is annotated with a logic category.
These pattern names and category labels serve as the primary structured signals for gold-skill construction.
LogicBench does not provide intermediate reasoning steps, formal derivations, or worked explanations; it contains only the problem statement and the final answer.
Each instance is associated with exactly one logic pattern, and each pattern is mapped to one gold skill.

% --- A.x.2 LLM Draft Generation ---
\subsubsection{LLM Draft Generation}
\label{sec:appendix-logicbench-draft}

For each pattern, the LLM receives the pattern name, its logic category, all associated questions and answers, and relevant Stanford Encyclopedia of Philosophy\footnote{\url{https://plato.stanford.edu}, a comprehensive, peer-reviewed reference for logic and philosophy.} pages as input.
Because the dataset provides no worked solutions, the LLM must synthesize the application procedure from its own knowledge of the inference rule, using the questions to understand the scope of application and the reference pages for canonical definitions.
The following prompt template is used:

\begin{promptbox}[Prompt Template for LogicBench]
\small\ttfamily
You are a logic education expert. Given an inference rule and a set of problems that test it, write a reusable skill document that teaches how to apply this rule to solve new problems.

\medskip
\textbf{Inference Rule:} \{pattern\_name\} \\
\textbf{Logic Category:} \{logic\_category\}

\textbf{Reference material:} \\
\{reference\_pages\}

\textbf{Problems testing this rule} (each with its ground truth answer): \\
\{questions\_and\_answers\}

\medskip
Based on the above materials, write a skill document with the following structure:
\begin{enumerate}
    \item \textbf{name}: Canonical inference rule name.
    \item \textbf{description}: One sentence summarizing what the rule states and its scope.
    \item \textbf{Content} (Markdown):
    \begin{itemize}
        \item Formal statement of the inference rule.
        \item When to recognize that this rule applies.
        \item Step-by-step application procedure.
        \item Common errors or pitfalls.
        \item One worked example using a new problem (not from the problems above).
    \end{itemize}
\end{enumerate}
\end{promptbox}

\subsubsection{Expert Revision}
\label{sec:appendix-logicbench-revision}

Beyond the general revision principles described in \S\ref{sec:golden_skill_construction}, LogicBench requires two dataset-specific refinements:
\begin{itemize}[leftmargin=*]
    \item \textbf{Narrative premise extraction}: In LogicBench, logical premises are often embedded in natural-language narratives that include scene-setting details, hedged expressions (e.g., ``typically,'' ``often,'' ``likely''), and other distractor content. While LLM drafts usually capture the target inference rule itself, they often under-specify how to identify the premises that are logically operative in the task. The expert therefore adds structured procedures for premise extraction, including how to separate reasoning-relevant premises from background context and how to interpret qualified conditionals in a rule-like manner.
    
    \item \textbf{Related fallacy disambiguation}: LogicBench also contains reasoning patterns that are easily confused with closely related fallacies (e.g., modus tollens vs.\ denying the antecedent, modus ponens vs.\ affirming the consequent). Although LLM drafts describe the target rule, they often do not explicitly contrast it with these neighboring invalid forms. The expert therefore adds distinction sections with recognition criteria to reduce confusion and prevent misapplication.
\end{itemize}

\subsubsection{Construction Examples}
\label{sec:appendix-logicbench-examples}

We present two complete construction examples from different logic categories.

\paragraph{Example 1: Modus Tollens.}
Modus tollens is a fundamental deductive rule appearing in both propositional and first-order logic; the key challenge lies in correctly extracting premises from narrative-rich problem statements.

\begin{rawbox}[Source Material]
\textbf{Pattern:} Modus Tollens \\
\textbf{Logic Category:} Propositional Logic, First-Order Logic \\
2 of 40 associated instances:

\medskip
\textbf{Q:} Daniel woke up on a gloomy morning and realized that he had forgotten to take his umbrella. Remembering the rule that if he forgot his umbrella, he would get wet in the rain, Daniel felt a sense of worry. However, Daniel managed to stay dry. Does this infer that Daniel forgot to take his umbrella? \\
\textbf{A:} no \quad (BQA)

\textbf{Q:} Isabella had been working hard on her project, knowing that if she finished it on time, she would receive a bonus. However, Isabella realized that she wouldn't be receiving a bonus. Does this entail that Isabella didn't finish her project on time? \\
\textbf{A:} yes \quad (BQA)
\end{rawbox}

\begin{draftbox}[LLM Drafted Skill]
\texttt{name:} Modus Tollens \\
\texttt{description:} Derive that the antecedent is false from a conditional and the denial of its consequent. \\[2pt]
\tcbline

\textbf{Modus Tollens}

Modus tollens (denying the consequent) derives a negative conclusion by contrapositive reasoning:

\textbf{Form:} If P, then Q. Q is false. Therefore, P is false. \\
Symbolic: $P \to Q,\; \neg Q \;\vdash\; \neg P$

\textbf{Application Procedure:}
\begin{enumerate}
    \item Identify the conditional premise: extract antecedent A and consequent C.
    \item Confirm that C is negated in the premises.
    \item Conclude $\neg$A. The antecedent must be false.
\end{enumerate}

\hfill \textcolor{red}{$\triangleright$ \textit{No guidance on narrative context or related fallacies}}

\textbf{Worked Example:}
If someone forgets their umbrella, they get wet in the rain. The person stayed dry. Therefore, they did not forget their umbrella. \hfill \textcolor{red}{$\triangleright$ \textit{Same scenario as an input instance}}

[\ldots]
\end{draftbox}

\begin{skillbox}[Golden Skill]
\texttt{name:} Modus Tollens \\
\texttt{description:} Denying the consequent: if P implies Q and Q is false, then P is false. Covers propositional and first-order logic. \\[2pt]
\tcbline

\textbf{Modus Tollens}

Modus tollens is the valid argument form: $P \to Q,\; \neg Q \;\vdash\; \neg P$.

\textbf{Application Procedure:}
\begin{enumerate}
    \item \textbf{Extract the conditional}: Identify A (antecedent) and C (consequent) from ``if A then C.''
    \item \textbf{Check the second premise}: Confirm that C is negated---``not C,'' ``C did not occur,'' or equivalent.
    \item \textbf{Apply modus tollens}: Conclude $\neg$A.
    \item \textbf{Answer the question}: Compare the question's target against $\neg$A.
\end{enumerate}

\textbf{Common Errors:}
\begin{itemize}
    \item \emph{Narrative context confusion.} A problem may describe A as having happened (scene-setting), then state $\neg$C. The narrative mention is not a logical premise---the operative premises are the conditional rule and the denial of C.
    \item \emph{Affirming the consequent (invalid):} ``If A then C; C; therefore A'' does not follow.
    \item \emph{Denying the antecedent (invalid):} ``If A then C; $\neg$A; therefore $\neg$C'' does not follow. Modus tollens requires negation of the \emph{consequent}, not the antecedent.
\end{itemize}

\textbf{Worked Example:}
``If a server authenticates successfully, encrypted data transfer begins.'' ``Encrypted transfer has not begun.'' \\
A = successful authentication, C = encrypted transfer. $\neg$C confirmed $\to$ $\neg$A: the server did not authenticate. \\
BQA: ``Did the server authenticate?'' $\to$ \textbf{no}.
\end{skillbox}

\noindent\textbf{Revision summary.}
(1)~The LLM draft correctly formalizes the inference rule but does not address the challenge of extracting premises from narrative scenarios.
LogicBench problems often embed the antecedent in scene-setting narrative (e.g., ``Daniel realized that he had forgotten his umbrella''), creating an apparent contradiction with the modus tollens conclusion ($\neg$A).
The expert added narrative context guidance and contrasted modus tollens with the commonly confused invalid forms (denying the antecedent, affirming the consequent).
(2)~The draft's worked example reuses the umbrella scenario from an input instance.
The expert replaced it with a newly constructed server authentication example.

% -------------------------------------------------------------------
\paragraph{Example 2: Default Reasoning with Irrelevant Information.}
This non-monotonic reasoning pattern tests whether the model can filter out logically irrelevant premises that do not affect the application of the default rule.

\begin{rawbox}[Source Material]
\textbf{Pattern:} Default Reasoning with Irrelevant Information \\
\textbf{Logic Category:} Non-Monotonic Logic \\
2 of 40 associated instances:

\medskip
\textbf{Q:} Snowflakes and crystals, which are typically white, often give the impression of being soft. However, snowflakes are not actually soft. Most white objects tend to possess a soft texture. Furthermore, snowflakes have a distinct six-sided structure. Does this entail that crystals are soft? \\
\textbf{A:} yes \quad (BQA)

\textbf{Q:} Pens and pencils are commonly used in classrooms. However, pens are not used in classrooms. Pens come in different colors. What conclusion would be most appropriate? \\
choice\_1: whiteboards are commonly used \;\; choice\_2: markers are preferred over pens \\
choice\_3: pencils are frequently used in classrooms \;\; choice\_4: pencils are used on the blackboard \\
\textbf{A:} choice\_3 \quad (MCQA)
\end{rawbox}

\begin{draftbox}[LLM Drafted Skill]
\texttt{name:} Default Reasoning with Irrelevant Information \\
\texttt{description:} Apply default rules correctly while filtering out irrelevant premises. \\[2pt]
\tcbline

\textbf{Default Reasoning with Irrelevant Information}

In default reasoning, a general rule applies to all members of a class unless a specific exception is stated. Additional information that does not modify the exception should not affect the conclusion.

\textbf{Application Procedure:}
\begin{enumerate}
    \item Identify the default rule and the exception.
    \item Apply the default to non-exception members.
\end{enumerate}

\hfill \textcolor{red}{$\triangleright$ \textit{No structured procedure for identifying and filtering distractors}}

\textbf{Worked Example:}
Snowflakes and crystals are typically soft. Snowflakes are an exception. Snowflakes have six sides. Are crystals soft? Yes---the six-sided shape is irrelevant. \hfill \textcolor{red}{$\triangleright$ \textit{Same entities as an input instance}}

[\ldots]
\end{draftbox}

\begin{skillbox}[Golden Skill]
\texttt{name:} Default Reasoning with Irrelevant Information \\
\texttt{description:} Applying a default rule correctly by filtering out logically irrelevant premises that do not affect the scope of the default. \\[2pt]
\tcbline

\textbf{Default Reasoning with Irrelevant Information}

In default reasoning, a premise set may contain information that is logically irrelevant to the conclusion. The ability to filter such distractors is as important as applying the rule itself.

\textbf{Step-by-Step Procedure:}
\begin{enumerate}
    \item Identify the default rule---look for ``typically,'' ``normally,'' ``usually.''
    \item Identify the exception---which individual is flagged as deviating from the default?
    \item \textbf{Actively set aside all additional information} that does not modify the default or re-designate any individual as an exception.
    \item Apply the default rule to the questioned individual, who remains a non-exceptional member.
    \item Conclude accordingly.
\end{enumerate}

\textbf{Why Distractors Mislead:}
\begin{itemize}
    \item \emph{Relevance inflation}: assuming every provided fact must factor into the conclusion.
    \item \emph{Narrative completion}: letting the overall story shade the conclusion beyond what the logical structure supports.
\end{itemize}

\textbf{Worked Example:}
``Birds typically can fly. Penguins cannot fly. Penguins live in Antarctica. Can eagles fly?'' \\
Step 1: Default = birds can fly. Step 2: Exception = penguins. Step 3: ``live in Antarctica'' is irrelevant---set aside. Step 4: Eagles are non-exceptional birds. Step 5: \textbf{yes}.
\end{skillbox}

\noindent\textbf{Revision summary.}
(1)~The LLM draft correctly states the default reasoning principle. Still, it reduces the procedure to two generic steps, without teaching how to \emph{identify} and filter distractor premises---the core challenge of this pattern.
The expert restructured the skill around an explicit five-step filtering procedure and added a ``Why Distractors Mislead'' section explaining the cognitive biases (relevance inflation, narrative completion) that cause models to incorporate irrelevant facts.
(2)~The draft's worked example reuses entities from an input instance (snowflakes, crystals, six-sided structure).
The expert replaced it with a newly constructed example.

% Appendix: ToolQA Construction Details

\newpage

\subsection{ToolQA}
\label{sec:appendix-toolqa}

The original ToolQA benchmark covers eight data domains. We exclude GSM8K, which primarily evaluates mathematical reasoning rather than tool-augmented question answering over an external reference corpus, and retain the remaining seven domains, resulting in 1{,}430 instances. Each instance poses a natural-language question grounded in one domain and requires an agent to interact with domain-specific external tools to retrieve, manipulate, or compute the answer. ToolQA provides a ground-truth \emph{answer} string for each instance, but does not include solution traces, worked explanations, or reusable skill annotations.
In our SRA-Bench, we define 14 annotation groups by crossing the seven retained domains with two difficulty levels. Easy questions generally require extracting a single piece of information from the external corpus, whereas hard questions require more compositional tool use, such as aggregation, comparison, filtering over multiple records, or derived computation. Each annotation group is represented by one gold skill.

The retained domains are associated with three types of tool-use capabilities. The tabular domains, including flight, coffee, Airbnb, and Yelp, require structured database operations via \texttt{LoadDB}, \texttt{FilterDB}, and \texttt{GetValue}. The text domains, Agenda and Scirex, require semantic retrieval over textual corpora through \texttt{RetrieveAgenda} and \texttt{RetrieveScirex}. The graph domain, DBLP, requires citation-network reasoning through \texttt{LoadGraph}, \texttt{NodeCheck}, \texttt{NeighbourCheck}, and \texttt{EdgeCheck}.
All retained domains can additionally use general-purpose tools, including \texttt{Calculate}, \texttt{PythonInterpreter}, and \texttt{SQLInterpreter}.

Crucially, the ToolQA repository publishes dataset generation code containing template-based question patterns for each domain$\times$difficulty.
Each template specifies a natural-language question pattern, the target data fields, and the answer computation procedure.
These templates, together with tool specifications and data schemas, provide the annotation-level context for skill construction and constitute construction-time knowledge beyond what the inference-time agent observes from the task prompt alone.
Each instance maps to exactly one domain $\times$ difficulty annotation, and each annotation is used to construct one gold skill.

% --- A.x.2 LLM Draft Generation ---
\subsubsection{LLM Draft Generation}
\label{sec:appendix-toolqa-draft}

For each domain $\times$ difficulty annotation, the LLM receives the domain name, difficulty level, tool specifications with signatures, data schema, question-generation templates from the published dataset generation code, and all associated questions and answers as input.
The question templates are particularly important because they reveal the systematic mapping from natural-language question patterns to data operations and answer formats.
This information is not available to the inference-time agent from the task prompt alone, but is useful for writing procedural guidance that generalizes across instances.
The following prompt template is used:

\begin{promptbox}[Prompt Template for ToolQA]
\small\ttfamily
You are a data analysis expert. Given a data domain, available tools, data schema, question templates, and QA examples, write a reusable skill document that teaches how to answer questions about this domain using the provided tools.

\medskip
\textbf{Domain:} \{domain\_name\} \quad \textbf{Difficulty:} \{difficulty\}

\textbf{Available tools:} 
\{tool\_specifications\}

\textbf{Data schema:} 
\{schema\}

\textbf{Question templates} (from dataset generation code): 
\{question\_templates\}

\textbf{QA pairs:} 
\{questions\_and\_answers\}

\medskip
Based on the above materials, write a skill document with the following structure:
\begin{enumerate}
    \item \textbf{name}: Domain and query type.
    \item \textbf{description}: One sentence summarizing the skill's scope.
    \item \textbf{Content} (Markdown):
    \begin{itemize}
        \item Data schema overview with key fields.
        \item Step-by-step tool workflow for each question type.
        \item Answer format conventions.
        \item Common errors or pitfalls.
    \end{itemize}
\end{enumerate}
\end{promptbox}

% --- A.x.3 Expert Revision ---
\subsubsection{Expert Revision}
\label{sec:appendix-toolqa-revision}

For ToolQA, expert revision focuses on making the drafted skills operationally usable by an inference-time agent. 
Although the agent is provided with tool descriptions and general-purpose few-shot ReAct demonstrations, these resources do not specify how to chain tools for different question patterns. 
The initial LLM drafts can often identify the relevant columns or fields from the question templates, but they rarely provide executable workflows, such as when a query should be handled by \texttt{FilterDB} versus \texttt{SQLInterpreter}, or when document retrieval should be followed by \texttt{PythonInterpreter} for aggregation beyond the retriever's top-$3$ results. 
Therefore, experts revise the skills by adding question-type-specific workflows that describe the intended sequence of tool calls, the required arguments, and the decision points for switching between tools.

Experts also add guidance on domain-specific data representations that are not explicitly documented in the original tool descriptions or schemas. 
For example, flight times are stored as HHMM-style numeric values (e.g., \texttt{2143.0} denotes 21:43), flight identifiers concatenate carrier codes and flight numbers (e.g., \texttt{AA2319} corresponds to carrier \texttt{AA} and flight number \texttt{2319}), and date formats may differ between natural-language questions and textual corpora. 
Therefore, the revised skills include lightweight decoding and normalization instructions that cover how to interpret encoded values, convert between formats, and construct valid tool inputs before execution.

% --- A.x.4 Construction Examples ---
\subsubsection{Construction Examples}
\label{sec:appendix-toolqa-examples}

% -------------------------------------------------------------------
\textbf{Example 1: Flights (Single-Record Lookup).}
% The flight-easy skill covers single-record lookups in the US domestic flight database; the key challenges are chaining the table tools into a structured workflow and correctly interpreting the HHMM time storage format.

\begin{rawbox}[Source Material]
\textbf{Domain:} flights (US domestic) \quad \textbf{Difficulty:} Easy \\
\textbf{Tools:} \texttt{LoadDB[flights]}, \texttt{FilterDB[cond, \ldots]}, \texttt{GetValue[column]} \\
\textbf{Key schema fields:} FlightDate, Origin, Dest, IATA\_Code\_Marketing\_Airline, Flight\_Number\_Marketing\_Airline, DepTime, ArrTime, AirTime, TaxiIn, Cancelled, \ldots \\
\textbf{Question templates} (3 of 10, from dataset generation code):
\begin{itemize}
    \item ``What was the departure time of \{carrier\}\{number\} from \{Origin\} to \{Dest\} on \{date\}?'' $\to$ DepTime
    \item ``How many minutes did \{carrier\}\{number\} take to taxi in on \{date\}?'' $\to$ TaxiIn
    \item ``Was \{carrier\}\{number\} from \{Origin\} to \{Dest\} cancelled on \{date\}?'' $\to$ Cancelled
\end{itemize}
2 of 100 associated instances:

\medskip
\textbf{Q:} What was the departure time of the AA2319 flight from MIA to LAS on 2022-06-05? 
\textbf{A:} 21:43

\textbf{Q:} How many minutes did the DL3538 flight take to taxi in on 2022-06-26? 
\textbf{A:} 3
\end{rawbox}

\begin{draftbox}[LLM Drafted Skill]
\texttt{name:} US Domestic Flight Data --- Easy Queries \\
\texttt{description:} Guide for answering single-record lookup questions about US domestic flight data. \\
% \tcbline

\textbf{Available Tools:}
\begin{itemize}
    \item \texttt{LoadDB[flights]}: Load the flight database.
    \item \texttt{FilterDB[cond1, cond2, \ldots]}: Filter rows matching conditions.
    \item \texttt{GetValue[column]}: Retrieve values from the filtered result.
\end{itemize}

\textbf{Question Types and Target Columns:}
\begin{itemize}
    \item Departure/arrival time $\to$ DepTime / ArrTime
    \item Flight duration $\to$ AirTime
    \item Taxi time $\to$ TaxiIn / TaxiOut
    \item Cancellation status $\to$ Cancelled
\end{itemize}

\hfill \textcolor{red}{$\triangleright$ \textit{Maps templates to columns but no step-by-step tool workflow or flight ID parsing}}

\textbf{Worked Example:}
Q: ``What was the departure time of the AA2319 flight from MIA to LAS on 2022-06-05?''
Filter by flight AA2319, origin MIA, destination LAS, date 2022-06-05. Get DepTime. \hfill \textcolor{red}{$\triangleright$ \textit{Same question; HHMM format not documented}}

\end{draftbox}

\begin{skillbox}[Golden Skill]
\texttt{name:} US Domestic Flight Data --- Single-Record Lookup Guide \\
\texttt{description:} Structured workflows for answering single-record lookup questions over the US domestic flight database, including schema reference, flight ID parsing, and time format interpretation. \\[2pt]
% \tcbline

\textbf{Flight Identifier Convention:}
A reference like \texttt{AA2319} encodes two separate columns: \texttt{IATA\_Code\_Marketing\_Airline=AA} and \texttt{Flight\_Number\_Marketing\_Airline=2319}. FilterDB requires them as separate conditions.

\textbf{Standard Lookup Workflow:}
\begin{enumerate}
    \item \texttt{LoadDB[flights]}
    \item \texttt{FilterDB[FlightDate=YYYY-MM-DD, Origin=XXX, IATA\_Code\_Marketing\_Airline=XX, Flight\_Number\_Marketing\_Airline=NNNN]}
    \item \texttt{GetValue[target\_column]}
\end{enumerate}
Use only the identifying fields given in the question; omitting unneeded fields is fine.

\textbf{Time Column Interpretation:}
Time columns (\texttt{DepTime}, \texttt{ArrTime}, etc.) store values as numeric HHMM without a colon---e.g., \texttt{2143.0} represents 21:43, \texttt{835.0} represents 08:35. Strip the \texttt{.0} suffix and insert a colon when reporting.

\textbf{When FilterDB Is Insufficient:}
FilterDB compares all values as strings (lexicographic), producing incorrect results for numeric columns. For numeric comparisons, use \texttt{SQLInterpreter} with \texttt{CAST(column AS REAL)}.

[\ldots]
\end{skillbox}

\noindent\textbf{Revision summary.}
(1)~The LLM draft correctly maps question types to target columns (using the templates) but does not specify the multi-step tool invocation workflow.
The draft says ``filter by flight AA2319'' without explaining that \texttt{AA2319} must be split into carrier code (\texttt{AA}) and flight number (\texttt{2319}) as separate FilterDB conditions---a parsing step that the agent must perform before any filter can succeed.
The expert added a structured three-step workflow (LoadDB $\to$ FilterDB $\to$ GetValue), the flight identifier convention, and guidance on when to switch from FilterDB to SQLInterpreter.
(2)~The draft does not document the HHMM time storage convention: time columns store numeric HHMM values (e.g., \texttt{2143.0} for 21:43), which the agent must decode to correctly interpret \texttt{GetValue} results and construct time-based queries.
(3)~The worked example reuses a question from an input instance.

% -------------------------------------------------------------------
\paragraph{Example 2: Agenda (Hard) --- Aggregation Queries.}
The agenda-hard skill covers population-level queries (counting events, finding free meeting slots) over a personal calendar corpus; the key challenge is recognizing that the semantic retrieval tool is fundamentally insufficient and switching to programmatic corpus processing.

\begin{rawbox}[Source Material]
\textbf{Domain:} agenda (personal calendars) \quad \textbf{Difficulty:} Hard \\
\textbf{Tools:} \texttt{RetrieveAgenda[query]} (returns top 3 passages), \texttt{PythonInterpreter[code]}, \texttt{Calculate[expr]} \\
\textbf{Corpus format:} JSONL; each event: ``On [date], [person] has [event] at [location] from [start] to [end].'' \\
\textbf{Question templates} (3 of 5, from dataset generation code):
\begin{itemize}
    \item ``How many events happen on \{date\}?'' $\to$ count all events on date
    \item ``How many people are unavailable between \{start\} and \{end\} on \{date\}?'' $\to$ count with time overlap
    \item ``When should I schedule a meeting with \{person\} on \{date\}?'' $\to$ comma-separated free slots
\end{itemize}
2 of 100 associated instances:

\medskip
\textbf{Q:} How many events happen on 2022/09/25 in the agenda table? \\
\textbf{A:} 24

\textbf{Q:} When should I schedule a meeting with Alice from 9:00 AM to 6:00 PM on 2022/01/06 in the agenda table? \\
\textbf{A:} 9:00 AM-1:00 PM, 3:00 PM-6:00 PM
\end{rawbox}

\begin{draftbox}[LLM Drafted Skill]
\texttt{name:} Agenda Event Data --- Hard Queries \\
\texttt{description:} Guide for answering aggregation and scheduling queries over the agenda event corpus. \\[2pt]
\tcbline

\textbf{Available Tools:}
\begin{itemize}
    \item \texttt{RetrieveAgenda[query]}: Search for relevant agenda events.
    \item \texttt{Calculate[expression]}: Perform arithmetic.
\end{itemize}

\textbf{Question Types:}
\begin{enumerate}
    \item Count events on a date $\to$ retrieve events, count results.
    \item Count unavailable people in a time window $\to$ retrieve events, count distinct people.
    \item Schedule a meeting $\to$ retrieve person's events, compute free slots.
\end{enumerate}

\hfill \textcolor{red}{$\triangleright$ \textit{RetrieveAgenda returns only top 3 passages---cannot count all 24 events on a date}}

\textbf{Worked Example:}
Q: ``How many events happen on 2022/09/25?''
\texttt{RetrieveAgenda[events 2022/09/25]} $\to$ count retrieved passages. \hfill \textcolor{red}{$\triangleright$ \textit{Same question; approach fundamentally wrong}}

[\ldots]
\end{draftbox}

\begin{skillbox}[Golden Skill]
\texttt{name:} Working with the Agenda Event Corpus: Aggregation and Population Queries \\
\texttt{description:} Programmatic corpus processing for agenda queries that require counting, time-overlap filtering, or free-slot scheduling beyond the retriever's top-3 limit. \\[2pt]
\tcbline

\textbf{Why RetrieveAgenda Cannot Answer Aggregation Queries:}
\texttt{RetrieveAgenda} returns only the top 3 passages by semantic similarity. This works for looking up a specific event but \emph{cannot} support queries that require counting or enumerating across all matching records. A date may have 15--40 events; retrieving 3 will produce the wrong count.

\textbf{Using PythonInterpreter for Corpus-Wide Queries:}
\begin{verbatim}
import json
corpus_path = "data/agenda/agenda_descriptions_merged.jsonl"
records = []
with open(corpus_path) as f:
    for line in f:
        if line.strip():
            records.append(json.loads(line))
\end{verbatim}

\textbf{Date Format Conversion:}
Questions use \texttt{YYYY/MM/DD}; corpus uses ``Month Day, Year'' (e.g., ``September 25, 2022''). Convert before filtering.

\textbf{Query Type $\to$ Tool Summary:}

\medskip
{\small
\begin{tabular}{ll}
\toprule
\textbf{Query type} & \textbf{Tool} \\
\midrule
Look up a specific event & \texttt{RetrieveAgenda} \\
Count events on a date & \texttt{PythonInterpreter} \\
Count people in time window & \texttt{PythonInterpreter} \\
Find free meeting slots & \texttt{PythonInterpreter} \\
\bottomrule
\end{tabular}
}

[\ldots]
\end{skillbox}

\noindent\textbf{Revision summary.}
(1)~The LLM draft suggests using \texttt{RetrieveAgenda} for all question types, not recognizing that aggregation queries (counting events, finding free slots) require access to the \emph{entire} corpus---not just the top 3 retrieved passages.
The expert restructured the skill around the fundamental distinction between single-event lookup (\texttt{RetrieveAgenda}) and corpus-wide queries (\texttt{PythonInterpreter}), adding complete code recipes for loading the corpus, converting date formats, filtering by time overlap, and computing free meeting slots.
(2)~The draft does not address the date format mismatch: questions use \texttt{YYYY/MM/DD} while the corpus stores dates as ``Month Day, Year'' (e.g., ``September 25, 2022''), requiring format conversion before records can be filtered programmatically.
(3)~The worked example reuses a question from an input instance.

% Appendix: MedCalc-Bench Construction Details
\newpage
\subsection{MedCalc-Bench}
\label{sec:appendix-medcalcbench}

% --- A.x.1 Source Material ---
% \subsubsection{Source Material}
% \label{sec:appendix-medcalcbench-source}

For MedCalc-Bench, we use the following fields from each instance: a \emph{patient note} describing the clinical vignette, a \emph{question} requesting the computation of a target clinical value, the \emph{calculator name}, the \emph{ground-truth answer}, and a \emph{ground-truth explanation}.
The ground-truth explanation provides a step-by-step walkthrough that extracts relevant values from the patient note and applies the calculator's formula or scoring rules.
The evaluation split contains 55 unique calculator names, with 20 test instances for each calculator.
Each instance is associated with exactly one calculator, and we construct one gold skill for each calculator.

For skill construction, the ground-truth explanations serve as the primary source material because they explicitly identify the relevant clinical variables and demonstrate the calculation procedure.
Compared with raw patient notes, these explanations provide a more information-dense basis for abstracting a reusable calculation skill.
The calculator name provides the structured annotation that links each instance to its corresponding gold skill.

% --- A.x.2 LLM Draft Generation ---
\subsubsection{LLM Draft Generation}
\label{sec:appendix-medcalcbench-draft}

For each calculator, the LLM receives the calculator name, all 20 ground-truth explanations, and the corresponding MDCalc\footnote{\url{https://www.mdcalc.com}, a widely used medical calculator database that provides formulas, scoring criteria, and evidence summaries for clinical calculators.} reference page as input.
The explanations provide worked examples from which the LLM abstracts a general procedure, while the MDCalc page provides the canonical formula specification and boundary conditions.
The following prompt template is used:

\begin{promptbox}[Prompt Template for MedCalc-Bench]
\small\ttfamily
You are a medical knowledge engineer. Given a set of worked calculation examples for a clinical calculator, write a reusable skill document that teaches how to apply this calculator to any patient.

\medskip
\textbf{Calculator name:} \{calculator\_name\}

\textbf{Reference page:} \{mdcalc\_page\}

\textbf{Worked examples} (20 patients, each showing extracted values and step-by-step calculation):\\
\{explanations\}

\medskip
Based on the above materials, write a skill document with the following structure:
\begin{enumerate}
    \item \textbf{name}: Canonical calculator name (include standard abbreviation if applicable).
    \item \textbf{description}: One sentence---what the calculator computes and its clinical use.
    \item \textbf{Content} (Markdown):
    \begin{itemize}
        \item Brief clinical context (1--2 sentences).
        \item Required inputs with expected units and common unit conversions.
        \item Step-by-step computation procedure using symbolic variables.
        \item A Python function \texttt{compute\_\{name\}(...)} implementing the calculator.
        \item One worked example using a new hypothetical patient (not from the examples above).
    \end{itemize}
\end{enumerate}
\end{promptbox}

% --- A.x.3 Expert Revision ---
\subsubsection{Expert Revision}
\label{sec:appendix-medcalcbench-revision}

Beyond the general revision criteria, expert revision for MedCalc-Bench focuses on making each skill both logically complete and executable. Since instance-level explanations usually reveal only the branch taken by a particular patient case, LLM-generated drafts may miss rare or unobserved branches, such as dialysis-dependent creatinine overrides or BMI-specific handling for underweight patients. The expert thus cross-checks each drafted skill against all annotated instances and the original MDCalc reference, completing the missing conditional logic in both the textual procedure and the accompanying Python implementation. The expert also verifies the executable tool by checking its outputs and correcting implementation details such as clamping rules, rounding behavior, and boundary conditions, ensuring that the final skill is operationally reliable rather than merely descriptive.

% --- A.x.4 Construction Examples ---
\subsubsection{Construction Examples}
\label{sec:appendix-medcalcbench-examples}

We present two construction examples illustrating how an LLM draft is revised into a finalized gold skill.
% -------------------------------------------------------------------
\paragraph{Example 1: MELD Na (UNOS/OPTN).}
The MELD Na score illustrates a complex calculator with multi-step conditional logic.

\begin{rawbox}[Source Material]
\textbf{Calculator:} MELD Na (UNOS/OPTN) \\
\textbf{Ground Truth Explanation} (1 of 20 instances):
The formula for computing the MELD Na is to first apply the following equation: MELD(i) = 0.957 $\times$ ln(Cr) + 0.378 $\times$ ln(bilirubin) + 1.120 $\times$ ln(INR) + 0.643. If the MELD(i) is greater than 11 after rounding to the nearest tenth and multiplying the MELD(i) by 10, we apply the following equation: MELD = MELD(i) + 1.32 $\times$ (137 $-$ Na) $-$ [0.033 $\times$ MELD(i) $\times$ (137 $-$ Na)]. The MELD Na score is capped at 40. The concentration of Na is mEq/L, the concentration of bilirubin is mg/dL, and the concentration of creatinine is mg/dL. If the patient's Na concentration is less than 125 mEq/L, we set it to 125 mEq/L and if the patient's the Na concentration is greater 137 mEq/L, we round it to 137 mEq/L. The concentration of creatinine is 2.0 mg/dL. The patient has not went through dialysis at least twice in the past week. [\ldots] The patient's bilirubin concentration is less than 1.0 mg/dL, and so we set the bilirubin concentration to 1.0 mg/dL. The patient's INR is 1.59. The sodium concentration is greater than 137 mEq/L, and so we set the sodium concentration to 137 mEq/L. Applying the first equation gives us 0.957 $\times$ ln(2.0) + 0.378 $\times$ ln(1.0) + 1.120 $\times$ ln(1.59) + 0.643 = 1.8257. Rounding to the nearest tenth makes the MELD(i) score 1.8. We then multiply by 10, making the MELD(i) score 18. Because the MELD(i) score is greater than 11, we then apply the second equation, giving us 18 + 1.32 $\times$ (137 $-$ 137) $-$ [0.033 $\times$ 1.8257 $\times$ (137 $-$ 137)] = 18. The MELD Na score is less than 40, and so we keep the score as it is. The patient's MELDNa score, rounded to the nearest integer, is \textbf{18 points}.
\end{rawbox}

\begin{draftbox}[LLM Drafted Skill]
\texttt{name:} MELD Na (UNOS/OPTN) \\
\texttt{description:} Calculate the MELD Na score to assess end-stage liver disease severity for transplant prioritization. \\[2pt]
\tcbline
\textbf{MELD Na (UNOS/OPTN)}

The MELD Na score estimates 90-day mortality risk in patients with end-stage liver disease and is used for liver transplant prioritization.

\textbf{Required Inputs:} Creatinine (mg/dL), bilirubin (mg/dL), INR, sodium (mEq/L), dialysis status.

\textbf{Step 1 --- Value clamping:}
Bilirubin: if $<$ 1.0, set to 1.0. INR: if $<$ 1.0, set to 1.0. Creatinine: clamp to [1.0, 3.0]. Sodium: clamp to [125, 137]. \hfill \textcolor{red}{$\triangleright$ \textit{Missing: dialysis $\to$ Cr\,=\,3.0}}

\textbf{Step 2 --- Compute MELD(i):}
$\text{MELD(i)} = 0.957 \ln(\text{Cr}) + 0.378 \ln(\text{Bili}) + 1.120 \ln(\text{INR}) + 0.643$

\textbf{Step 3 --- Sodium adjustment:}
$\text{MELD Na} = \text{MELD(i)}{\times}10 + 1.32 (137 {-} \text{Na}) - 0.033 \cdot \text{MELD(i)}{\times}10 \cdot (137 {-} \text{Na})$

Cap at 40 and round to nearest integer. \hfill \textcolor{red}{$\triangleright$ \textit{Missing: MELD(i)\,$\leq$\,11 bypass}}

\textbf{Tool:}
\begin{verbatim}
def compute_meld_na(creatinine, bilirubin, inr, sodium, on_dialysis):
    if bilirubin < 1.0: bilirubin = 1.0
    if inr < 1.0: inr = 1.0
    creatinine = max(1.0, min(creatinine, 3.0))  # no dialysis handling
    sodium = max(125, min(sodium, 137))
    meld_i = (0.957 * math.log(creatinine) + 0.378 * math.log(bilirubin)
              + 1.120 * math.log(inr) + 0.643)
    meld_i_scaled = round(round(meld_i, 1) * 10)
    meld_na = (meld_i_scaled + 1.32 * (137 - sodium)
               - 0.033 * meld_i_scaled * (137 - sodium))  # always applied
    return round(min(meld_na, 40))
\end{verbatim}

[\ldots]
\end{draftbox}

\begin{skillbox}[Golden Skill]
\texttt{name:} MELD Na (UNOS/OPTN) \\
\texttt{description:} Compute MELD Na score for liver transplant prioritization from creatinine, bilirubin, INR, and sodium. \\[2pt]
\tcbline

\textbf{MELD Na (UNOS/OPTN)}

The MELD Na score (Model for End-Stage Liver Disease with Sodium) is used to prioritize patients on the liver transplant waiting list. It extends the original MELD score by incorporating serum sodium to better predict 90-day mortality in end-stage liver disease.

\textbf{Required Inputs:} Creatinine (mg/dL), bilirubin (mg/dL), INR, sodium (mEq/L), dialysis status.

\textbf{Step 1 --- Value clamping before MELD(i):}
\begin{itemize}
    \item Creatinine: if on dialysis ($\geq$2$\times$/week), set to 3.0. Otherwise clamp to [1.0, 3.0].
    \item Bilirubin: if $<$ 1.0, set to 1.0.
    \item INR: if $<$ 1.0, set to 1.0.
\end{itemize}

\textbf{Step 2 --- Compute MELD(i):}
$$\text{MELD(i)} = 0.957 \ln(\text{Cr}) + 0.378 \ln(\text{Bili}) + 1.120 \ln(\text{INR}) + 0.643$$

\textbf{Step 3 --- Check sodium adjustment:}
Round MELD(i) to nearest tenth, multiply by 10. If $\leq 11$: final score = MELD(i) $\times$ 10. If $> 11$: proceed to Step 4.

\textbf{Step 4 --- Sodium clamping:} Clamp Na to [125, 137] mEq/L.

\textbf{Step 5 --- Sodium adjustment:}
$$\text{MELD Na} = \text{MELD(i)}{\times}10 + 1.32 (137 {-} \text{Na}) - 0.033 \cdot \text{MELD(i)}{\times}10 \cdot (137 {-} \text{Na})$$

\textbf{Step 6 --- Cap:} If MELD Na $>$ 40, set to 40. Round to nearest integer.

\textbf{Tool:}
\begin{verbatim}
def compute_meld_na(creatinine, bilirubin, inr, sodium, on_dialysis):
    if on_dialysis: creatinine = 3.0
    else: creatinine = max(1.0, min(creatinine, 3.0))
    if bilirubin < 1.0: bilirubin = 1.0
    if inr < 1.0: inr = 1.0
    meld_i = (0.957 * math.log(creatinine) + 0.378 * math.log(bilirubin)
              + 1.120 * math.log(inr) + 0.643)
    meld_i_scaled = round(round(meld_i, 1) * 10)
    sodium = max(125, min(sodium, 137))
    if meld_i_scaled > 11:
        meld_na = (meld_i_scaled + 1.32 * (137 - sodium)
                   - 0.033 * meld_i_scaled * (137 - sodium))
    else:
        meld_na = meld_i_scaled
    return round(min(meld_na, 40))
\end{verbatim}

\textbf{Worked Example:}
Patient: 58-year-old man with cirrhosis. Cr 1.8, Bili 3.5, INR 1.4, Na 130. Not on dialysis.

Step 1: All values in range, no clamping needed.

Step 2--3: \texttt{TOOL\_CALL: compute\_meld\_na(1.8, 3.5, 1.4, 130, False) $\to$ 25}

The patient's MELD Na score is \textbf{25}.
\end{skillbox}

\noindent\textbf{Revision summary.}
The LLM draft captures the correct formulas and clamping rules for the majority of patients, but omits two conditional branches---in both the textual procedure and the Python tool.
(1)~It omits the dialysis-dependent creatinine override (Cr $\to$ 3.0 when on dialysis $\geq$2$\times$/week), a branch that appears in only 2 of the 20 input instances.
(2)~It always applies the sodium adjustment, missing the MELD(i) $\leq$ 11 threshold that bypasses it entirely---a condition that none of the 20 instances happen to trigger.
The expert added both branches by consulting the MDCalc reference page and restructured the procedure into 6 explicit steps.

% -------------------------------------------------------------------

\newpage
\paragraph{Example 2: Creatinine Clearance (Cockcroft-Gault Equation).}
The Cockcroft-Gault equation appears to be a single formula, but the BMI-based weight adjustment makes it procedurally complex.

\begin{rawbox}[Source Material]
\textbf{Calculator:} Creatinine Clearance (Cockcroft-Gault Equation) \\
\textbf{Ground Truth Explanation} (1 of 20 instances, abbreviated):
The formula for computing Cockcroft-Gault is given by CrCl = ((140 $-$ age) $\times$ weight $\times$ gender\_coefficient) / (serum creatinine $\times$ 72), where the gender\_coefficient is 1 if male, and 0.85 if female. The serum creatinine concentration is in mg/dL.
The patient's gender is female, which means that the gender coefficient is 0.85. The patient is 62 years old.
The concentration of creatinine is 85.0 umol/L. We need to convert the concentration to mg/dL. Let's first convert the mass of creatinine from umol to mg. [\ldots 6-step molar mass conversion\ldots] This will result to 9.62 mg creatinine/10.0 dL = 0.962 mg creatinine/dL.
[\ldots] Because the patient is overweight/obese, we use the adjusted body weight formula. For females, the ideal body weight (IBW) is calculated as follows: 45.5 kg + 2.3 kg $\times$ (height (in inches) $-$ 60). [\ldots] ABW = IBW + 0.4 $\times$ (weight $-$ IBW) = 52.38 + 0.4 $\times$ (72.0 $-$ 52.38) = 60.23 kg. [\ldots]
Plugging the patient's values gives us ((140 $-$ 62) $\times$ 60.23 $\times$ 0.85) / (0.962 $\times$ 72) = \textbf{57.65 mL/min}.
\end{rawbox}

\begin{draftbox}[LLM Drafted Skill]
\texttt{name:} Cockcroft-Gault Creatinine Clearance \\
\texttt{description:} Estimate creatinine clearance using the Cockcroft-Gault equation for drug dosing in renal impairment. \\[2pt]
\tcbline

\textbf{Creatinine Clearance (Cockcroft-Gault)}

The Cockcroft-Gault equation estimates creatinine clearance (CrCl) to guide drug dosing in renal impairment.

\textbf{Required Inputs:}
Age (years), sex, height, weight, serum creatinine (mg/dL; from $\mu$mol/L: $\div\,88.42$).

\textbf{Step 1 --- Weight adjustment:}
Compute IBW: Male $= 50 + 2.3 \times (h_{\text{in}} - 60)$; Female $= 45.5 + 2.3 \times (h_{\text{in}} - 60)$.
For overweight/obese patients, use ABW $= \text{IBW} + 0.4 \times (\text{actual} - \text{IBW})$.

\hfill \textcolor{red}{$\triangleright$ \textit{Only overweight branch; missing underweight and normal-weight rules}}

\textbf{Step 2 --- Apply Cockcroft-Gault:}
$\text{CrCl} = \frac{(140 - \text{age}) \times w_{\text{adj}} \times g}{\text{Cr} \times 72}$ where $g = 1.0$ (male) or $0.85$ (female).

\textbf{Tool:}
\begin{verbatim}
def compute_cg_crcl(age, weight_kg, height_cm, creatinine, is_female):
    h_in = height_cm * 0.393701
    ibw = (45.5 if is_female else 50) + 2.3 * (h_in - 60)
    adj_w = ibw + 0.4 * (weight_kg - ibw)  # always uses ABW
    g = 0.85 if is_female else 1.0
    return ((140 - age) * adj_w * g) / (creatinine * 72)
\end{verbatim}
\hfill \textcolor{red}{$\triangleright$ \textit{No BMI check; always applies overweight formula}}

\textbf{Worked Example:}
Patient: 62-year-old female, 165 cm, 72.0 kg, Cr 0.962 mg/dL (from 85.0 $\mu$mol/L).
[\ldots] CrCl = \textbf{57.65 mL/min}. \hfill \textcolor{red}{$\triangleright$ \textit{Values from an input instance}}
\end{draftbox}

\begin{skillbox}[Golden Skill]
\texttt{name:} Creatinine Clearance (Cockcroft-Gault Equation) \\
\texttt{description:} Compute creatinine clearance using the Cockcroft-Gault equation with BMI-based weight adjustment. \\[2pt]
\tcbline

\textbf{Creatinine Clearance (Cockcroft-Gault Equation)}

The Cockcroft-Gault equation estimates creatinine clearance (CrCl) as a surrogate for GFR, widely used for drug dosing adjustments in renal impairment. It uses an adjusted body weight determined from the patient's BMI category.

\textbf{Required Inputs:}
Age (years), sex, height (cm; from inches: $\times\,2.54$), weight (kg; from lbs: $\times\,0.4536$), serum creatinine (mg/dL; from $\mu$mol/L: $\div\,88.42$).

\textbf{Step 1 --- Calculate BMI:}
$\text{BMI} = \text{weight\_kg} \;/\; (\text{height\_cm}/100)^2$

\textbf{Step 2 --- Determine adjusted body weight:}
\begin{itemize}
    \item Underweight (BMI $<$ 18.5): use actual body weight.
    \item Normal (18.5--24.9): use $\min(\text{IBW}, \text{actual weight})$.
    \item Overweight/obese ($\geq$ 25): use $\text{ABW} = \text{IBW} + 0.4 \times (\text{actual} - \text{IBW})$.
\end{itemize}
IBW: Male $= 50 + 2.3 \times (h_{\text{in}} - 60)$; \; Female $= 45.5 + 2.3 \times (h_{\text{in}} - 60)$.

\textbf{Step 3 --- Apply Cockcroft-Gault:}
$$\text{CrCl} = \frac{(140 - \text{age}) \times w_{\text{adj}} \times g}{\text{Cr} \times 72} \;\;\text{(mL/min)}$$
where $g = 1.0$ (male) or $0.85$ (female).

\textbf{Tool:}
\begin{verbatim}
def compute_cg_crcl(age, weight_kg, height_cm, creatinine, is_female):
    bmi = weight_kg / (height_cm / 100) ** 2
    h_in = height_cm * 0.393701
    ibw = (45.5 if is_female else 50) + 2.3 * (h_in - 60)
    g = 0.85 if is_female else 1.0
    if bmi < 18.5:
        adj_w = weight_kg
    elif bmi < 25:
        adj_w = min(ibw, weight_kg)
    else:
        adj_w = ibw + 0.4 * (weight_kg - ibw)
    return ((140 - age) * adj_w * g) / (creatinine * 72)
\end{verbatim}

\textbf{Worked Example:}
Patient: 65-year-old female, 160 cm, 55 kg, Cr 1.0 mg/dL.

Step 1: BMI = 55 / 1.60$^2$ = 21.5 $\to$ normal weight.

Step 2: IBW = 52.38 $<$ 55, so $w_{\text{adj}}$ = $\min(52.38, 55)$ = 52.38 kg.

Step 3: CrCl = $(140 - 65) \times 52.38 \times 0.85 \;/\; (1.0 \times 72)$ = \textbf{46.38 mL/min}.
\end{skillbox}

\vspace{10mm}

\noindent\textbf{Revision summary.}
(1)~The LLM draft only implements the overweight/obese ABW adjustment---the text omits underweight and normal-weight rules, and the Python tool always computes ABW regardless of BMI.
The expert added the full three-way branching (underweight $\to$ actual weight; normal $\to$ $\min(\text{IBW}, \text{actual})$; overweight $\to$ ABW) by consulting the MDCalc reference.
(2)~The draft's worked example directly reuses values from an input instance.
The expert replaced it with a newly constructed patient (65-year-old female, normal BMI) that exercises the $\min(\text{IBW}, \text{actual})$ branch.

% Appendix: CHAMP Construction Details
\newpage

\subsection{CHAMP}
\label{sec:appendix-champ}

Each instance in CHAMP contains a competition mathematics \emph{question}, a ground truth \emph{answer} (which may be a number, algebraic expression, yes/no, or brief phrase), and \emph{official solution steps} detailing the intended approach.
The dataset provides 89 unique concept annotations across eight mathematical domains: number theory (29 concepts), inequality (18), polynomial (18), algebra (9), combinatorics (8), sequence (3), trigonometry (3), and function (1).
Each concept is accompanied by a \emph{concept description}---a formal statement of the mathematical theorem or identity (e.g., ``For non-negative $x, y$: $(x{+}y)/2 \geq \sqrt{xy}$, with equality iff $x{=}y$'')---which serves as annotation-level context.

CHAMP uses \textbf{multi-label annotations}: each instance is tagged with all concepts relevant to its solution (average 1.7 per instance), and 49\% of instances require two or more concepts.
Each concept corresponds to one gold skill.

% --- A.x.2 LLM Draft Generation ---
\subsubsection{LLM Draft Generation}
\label{sec:appendix-champ-draft}

For each concept, the LLM receives the concept name, its mathematical definition, category, all associated questions with answers and official solution steps, and relevant AoPS Wiki\footnote{\url{https://artofproblemsolving.com/wiki}---a community-maintained mathematics reference with extensive coverage of competition problem-solving techniques.} pages as input.
The solution steps provide concrete worked examples of how the concept is applied, giving the LLM significantly more procedural context than in TheoremQA or LogicBench where no solution steps are available.
The following prompt template is used:

\begin{promptbox}[Prompt Template for CHAMP]
\small\ttfamily
You are a mathematics education expert. Given a mathematical concept and competition problems that apply it, write a reusable skill document that teaches how to apply this concept to solve new problems.

\medskip
\textbf{Concept:} \{concept\_name\} \\
\textbf{Category:} \{category\} \\
\textbf{Definition:} \{concept\_definition\}

\textbf{Reference material:} \\
\{reference\_pages\}

\textbf{Problems applying this concept} (each with its answer and official solution steps): \\
\{problems\_with\_solutions\}

\medskip
Based on the above materials, write a skill document with the following structure:
\begin{enumerate}
    \item \textbf{name}: Canonical concept name.
    \item \textbf{description}: One sentence summarizing what the concept computes or states.
    \item \textbf{Content} (Markdown):
    \begin{itemize}
        \item Statement of the concept (theorem or formula).
        \item When to recognize that this concept applies.
        \item Step-by-step application procedure.
        \item Common errors or pitfalls.
        \item One worked example using a new problem (not from the problems above).
    \end{itemize}
\end{enumerate}
\end{promptbox}

% --- A.x.3 Expert Revision ---
\subsubsection{Expert Revision}
\label{sec:appendix-champ-revision}

In addition to the general revision criteria, the following dataset-specific activities apply:
\begin{itemize}[leftmargin=*]
    \item \textbf{Recognition pattern development}: Competition math problems rarely name the concept they require.
    The concept definition tells \emph{what} a theorem states, and solution steps demonstrate specific applications, but neither systematically identifies the structural cues that signal applicability across diverse problem types.
    LLM drafts restate the theorem and walk through example calculations without abstracting recognition patterns.
    The expert adds problem-type classification and structural signatures that indicate when the concept should be applied.
    \item \textbf{Solution strategy systematization}: Official solution steps show the concept applied in ad-hoc fashion to individual problems, but competition math concepts often encompass multiple distinct application strategies (e.g., modular arithmetic for impossibility proofs vs.\ residue enumeration vs.\ periodicity analysis).
    LLM drafts present the concept as a single tool with one usage mode.
    The expert organizes the skill around a typology of problem types, each with its own systematic procedure.
\end{itemize}

% --- A.x.4 Construction Examples ---
\subsubsection{Construction Examples}
\label{sec:appendix-champ-examples}

We present two complete construction examples from different mathematical domains.

% -------------------------------------------------------------------
\paragraph{Example 1: AM-GM Inequality (Two Variables).}
The AM-GM inequality is a fundamental bounding tool; the key challenge lies in recognizing when an optimization problem can be solved by identifying the right pair of non-negative terms.

\begin{rawbox}[Source Material]
\textbf{Concept:} AM-GM Inequality (Two Variables) \\
\textbf{Category:} Inequality \\
\textbf{Definition:} For non-negative $x, y$: $(x{+}y)/2 \geq \sqrt{xy}$, with equality iff $x = y$. \\
2 of 6 associated instances:

\medskip
\textbf{Q:} If $a_i > 0$ for $i = 1, 2, \ldots, n$ and $a_1 a_2 \cdots a_n = 1$, what is the smallest value of $(1{+}a_1)(1{+}a_2)\cdots(1{+}a_n)$ as an expression of $n$? \\
\textbf{A:} $2^n$

\textbf{Q:} For positive numbers $a, b, c$, what is the smallest value of $(a{+}b)(b{+}c)(a{+}c)/(abc)$? \\
\textbf{A:} 8
\end{rawbox}

\begin{draftbox}[LLM Drafted Skill]
\texttt{name:} AM-GM Inequality (Two Variables) \\
\texttt{description:} Establish bounds on sums and products of non-negative numbers using the AM-GM inequality. \\[2pt]
\tcbline

\textbf{AM-GM Inequality (Two Variables)}

\textbf{Statement:} For non-negative $x, y$: $(x{+}y)/2 \geq \sqrt{xy}$, equivalently $x + y \geq 2\sqrt{xy}$. Equality holds iff $x = y$.

\textbf{Procedure:}
\begin{enumerate}
    \item Identify two non-negative terms $u, v$.
    \item Apply AM-GM: $u + v \geq 2\sqrt{uv}$.
    \item Verify that the equality condition $u = v$ is achievable.
\end{enumerate}

\hfill \textcolor{red}{$\triangleright$ \textit{No recognition patterns---when should one try AM-GM?}}

\textbf{Worked Example:}
For positive $a, b, c$: $(a{+}b)(b{+}c)(a{+}c)/(abc)$. Apply AM-GM to each factor: $a{+}b \geq 2\sqrt{ab}$, etc. Product $\geq 8$. \hfill \textcolor{red}{$\triangleright$ \textit{Same problem as an input instance}}

[\ldots]
\end{draftbox}

\begin{skillbox}[Golden Skill]
\texttt{name:} AM-GM Inequality (Two Variables) \\
\texttt{description:} Apply AM-GM for two variables to establish bounds on sums, products, and square root comparisons. \\[2pt]
\tcbline

\textbf{AM-GM Inequality (Two Variables)}

\textbf{Statement:} For non-negative $x, y$: $(x{+}y)/2 \geq \sqrt{xy}$. Equality holds iff $x = y$.

\textbf{Recognition:}
Use AM-GM when:
\begin{itemize}
    \item A sum $x{+}y$ needs a lower bound in terms of $\sqrt{xy}$, or a product needs an upper bound.
    \item The expression contains $t + 1/t$ for $t > 0$ (directly gives $t + 1/t \geq 2$).
    \item A product of factors $(a{+}b)(b{+}c)\cdots$ appears---apply AM-GM to each factor, then multiply.
\end{itemize}

\textbf{Procedure:}
\begin{enumerate}
    \item \textbf{Identify terms}: Find two non-negative terms $u, v$ whose product or sum appears in the expression.
    \item \textbf{Apply AM-GM}: $u + v \geq 2\sqrt{uv}$.
    \item \textbf{Chain or multiply}: To bound a product of factors, apply AM-GM to each factor and multiply the results.
    \item \textbf{Verify equality}: Confirm $u = v$ is achievable under the problem's constraints.
\end{enumerate}

\textbf{Common Pitfalls:}
\begin{itemize}
    \item \emph{Non-negativity}: both terms must be non-negative; AM-GM does not apply otherwise.
    \item \emph{Chained equality}: when AM-GM is applied multiple times, all equality conditions must hold simultaneously---check that they are mutually compatible.
\end{itemize}

\textbf{Worked Example:}
For $t > 0$, find the minimum of $t + 4/t$.
Apply AM-GM with $u = t$, $v = 4/t$: $t + 4/t \geq 2\sqrt{4} = 4$.
Equality when $t = 4/t$, i.e., $t = 2$. Minimum is $\mathbf{4}$.
\end{skillbox}

\noindent\textbf{Revision summary.}
(1)~The LLM draft correctly states the AM-GM formula and basic procedure but does not teach how to recognize when a problem calls for AM-GM.
Problems like ``minimize $(1{+}a_1)\cdots(1{+}a_n)$ given $\prod a_i = 1$'' require recognizing the per-factor AM-GM pattern, which is not obvious from the formula alone.
The expert added a ``Recognition'' section with structural cues (product-sum patterns, reciprocal expressions, products of factors).
(2)~The draft's worked example uses the same problem as an input instance.
The expert replaced it with a newly constructed $t + 4/t$ example that exercises the reciprocal-expression pattern.

% -------------------------------------------------------------------
\paragraph{Example 2: Modular Arithmetic.}
Modular arithmetic is the most frequently applied concept in CHAMP (32 instances); the main challenge is selecting the right modulus and application strategy from among several distinct problem types.

\begin{rawbox}[Source Material]
\textbf{Concept:} Modular Arithmetic \\
\textbf{Category:} Number Theory \\
\textbf{Definition:} $(a{+}b) \bmod m = ((a \bmod m) + (b \bmod m)) \bmod m$. $(a \cdot b) \bmod m = ((a \bmod m) \cdot (b \bmod m)) \bmod m$. $a^k \bmod m = ((a \bmod m)^k) \bmod m$. \\
2 of 32 associated instances:

\medskip
\textbf{Q:} Are there integer solutions to the equation $(x^2{-}1)(y^2{-}1)+1985=z^2$? \\
\textbf{A:} No

\textbf{Q:} Let $(x, y, z)$ be an integer-valued solution to $x^2+y^2=z^2$. At least how many of the three numbers must be divisible by 5? \\
\textbf{A:} At least 1 number
\end{rawbox}

\begin{draftbox}[LLM Drafted Skill]
\texttt{name:} Modular Arithmetic \\
\texttt{description:} Use modular arithmetic rules to analyze divisibility and residue patterns. \\[2pt]
\tcbline

\textbf{Modular Arithmetic}

\textbf{Core Rules:} $(a{+}b) \bmod m = ((a \bmod m) + (b \bmod m)) \bmod m$. Similar for multiplication and exponentiation.

\textbf{Procedure:}
\begin{enumerate}
    \item Choose a modulus $m$.
    \item Compute residues of each side of the equation.
    \item Check if the residues are compatible.
\end{enumerate}

\hfill \textcolor{red}{$\triangleright$ \textit{No guidance on modulus selection or problem-type classification}}

\textbf{Worked Example:}
Show $x^2{+}y^2{=}z^2$ requires at least one variable divisible by 5: check all residues mod~5. Squares mod 5 are $\{0, 1, 4\}$. If none is divisible by 5 \ldots \hfill \textcolor{red}{$\triangleright$ \textit{Same problem as an input instance}}

[\ldots]
\end{draftbox}

\begin{skillbox}[Golden Skill]
\texttt{name:} Modular Arithmetic: Residue Analysis \\
\texttt{description:} Apply modular arithmetic to prove impossibility, enumerate residues, analyze periodicities, and use CRT for simultaneous divisibility. \\[2pt]
\tcbline

\textbf{Modular Arithmetic: Residue Analysis}

\textbf{Problem Types and Recognition:}
\begin{itemize}
    \item \emph{Type 1---Impossibility}: ``Find all integer solutions'' $\to$ choose $m$ so residue sets of each side are disjoint.
    \item \emph{Type 2---Residue enumeration}: ``What are the possible values mod $m$?'' $\to$ compute $f(x) \bmod m$ for each residue class.
    \item \emph{Type 3---Divisibility}: ``Show $d \mid f(x)$ for all $x$'' $\to$ verify $f(x) \bmod d = 0$ across all residue classes.
    \item \emph{Type 4---Periodicity}: ``Find $f(n) \bmod m$ for large $n$'' $\to$ compute until a cycle appears.
\end{itemize}

\textbf{Modulus Selection:}
\begin{itemize}
    \item $m = 3, 9$: distinguish squares (0, 1 mod 3) and cubes.
    \item $m = 4, 8$: distinguish squares (0, 1, 4 mod 8), powers of 2.
    \item Start with the modulus suggested by the coefficients; try others if the first fails.
\end{itemize}

\textbf{Procedure:}
\begin{enumerate}
    \item Classify the problem type (impossibility, enumeration, divisibility, or periodicity).
    \item Select modulus based on problem structure and coefficient cues.
    \item Enumerate relevant residues (for all integers, or restricted to primes, etc.).
    \item Compare residue sets or compute the period as needed.
\end{enumerate}

\textbf{Worked Example:}
Prove that no integers satisfy $x^2 + y^2 + z^2 = 8k + 7$.
Type 1 (impossibility). Choose $m = 8$: $x^2 \bmod 8 \in \{0, 1, 4\}$.
Possible sums of three elements from $\{0, 1, 4\}$: $\{0, 1, 2, 3, 4, 5, 6\}$---7 is not achievable.
Since $8k + 7 \equiv 7 \pmod{8}$ is unreachable, no solutions exist.
\end{skillbox}

\noindent\textbf{Revision summary.}
(1)~The LLM draft states the modular arithmetic identities and a generic three-step procedure but does not distinguish among the different problem types or provide guidance on modulus selection---the core strategic decisions.
The expert organized the skill around a four-type classification (impossibility, enumeration, divisibility, periodicity) with type-specific procedures and added concrete modulus selection heuristics by consulting the AoPS Wiki.
(2)~The draft's worked example uses the same Pythagorean-triple divisibility problem from an input instance.
The expert replaced it with a newly constructed impossibility example ($x^2 + y^2 + z^2 \neq 8k + 7$) that demonstrates the full Type-1 workflow.

\newpage
\subsection{BigCodeBench}
\label{sec:appendix-bigcodebench}

Each instance in BigCodeBench specifies a Python function to implement. 
The task description defines the expected behavior, inputs and outputs, and, when applicable, error-handling requirements, together with a function signature and a set of relevant library/import requirements.
The dataset also provides a canonical solution and a unit-test suite for execution-based evaluation, where model performance is commonly measured by metrics such as \texttt{pass@1}.
BigCodeBench covers 139 Python libraries.

BigCodeBench provides multi-label library annotations: each instance is associated with the libraries used or permitted in the reference implementation, with an average of 2.8 libraries per task.
In our conversion, each distinct library is converted into one gold skill, and an instance may therefore be associated with multiple gold skills.
The dataset does not provide annotation-level context such as library descriptions, API summaries, or reusable usage guidance.

\subsubsection{LLM Draft Generation}
\label{sec:appendix-bigcodebench-draft}

For each library, the LLM receives the library name, the associated task descriptions and canonical solutions, and relevant official documentation\footnote{e.g., \url{https://docs.python.org/3/library/re.html}, \url{https://matplotlib.org/stable/api/}---official Python and third-party library API references.} pages as input.
The canonical solutions are used to identify concrete API usage patterns: which functions are invoked, how parameters are configured, what objects are returned, and how different calls are composed to satisfy task requirements.
This information is often critical for generating code that passes unit tests, but is not always explicitly stated in the natural-language task descriptions.
To avoid instance-level leakage, the final gold skills are revised to describe reusable API contracts and task-oriented recipes, rather than to copy benchmark-specific solutions.
The following prompt template is used:

\begin{promptbox}[Prompt Template for BigCodeBench]
\small\ttfamily
You are a Python programming expert. Given a library and a set of coding tasks that use it, write a reusable skill document that teaches how to use this library's API correctly in similar tasks.

\medskip
\textbf{Library:} \{library\_name\}

\textbf{Official documentation:} \\
\{doc\_pages\}

\textbf{Tasks using this library} (each with its task description and canonical solution): \\
\{tasks\_and\_solutions\}

\medskip
Based on the above materials, write a skill document with the following structure:
\begin{enumerate}
    \item \textbf{name}: Library name with a descriptive subtitle.
    \item \textbf{description}: One sentence summarizing the practical API patterns covered.
    \item \textbf{Content} (Markdown):
    \begin{itemize}
        \item Core API functions with calling conventions and return types.
        \item Common usage patterns with code examples.
        \item Common errors or pitfalls.
    \end{itemize}
\end{enumerate}
\end{promptbox}

% --- A.x.3 Expert Revision ---
\subsubsection{Expert Revision}
\label{sec:appendix-bigcodebench-revision}

For BigCodeBench, expert revision focuses on making each skill directly useful for code-generation tasks that are evaluated by automated unit tests. 
Since passing these tests often depends on exact API semantics, merely describing a library at a conceptual level is insufficient. 
For example, an LLM-generated draft may state that \texttt{re} supports pattern matching and substitution, but omit the behavioral details that determine whether the generated code is correct, such as return types, parameter-specific effects, and edge-case behaviors. 
Experts thus revise such drafts into precise API contracts, including concrete notes on return values, parameter interactions, and representative examples; for instance, \texttt{re.findall} returns captured group contents when the pattern contains a single capturing group, and \texttt{plt.hist()} returns a tuple rather than an Axes object.

In addition, experts reorganize the skills from function-centric descriptions into task-oriented recipes. 
A single library may support many distinct programming needs, such as URL removal, word tokenization, file matching, data validation, or visualization. 
Listing APIs one by one does not clearly indicate how to combine these functions for a particular task. 
The revised skills are therefore structured around reusable code patterns indexed by their intended use, enabling a model to identify the relevant recipe and apply it to the current programming problem.

% --- A.x.4 Construction Examples ---
\subsubsection{Construction Examples}
\label{sec:appendix-bigcodebench-examples}

We present two complete construction examples from different library domains.

% -------------------------------------------------------------------
\paragraph{Example 1: \texttt{re} (Regular Expressions).}
The \texttt{re} library is used across 139 instances for text processing tasks; the key challenge lies in the subtle behavioral differences between API functions---for example, \texttt{re.findall} with a capturing group returns group contents rather than full matches, and chained \texttt{re.sub} operations interact when later substitutions depend on earlier ones.

\begin{rawbox}[Source Material]
\textbf{Library:} \texttt{re} --- Regular Expressions \\
2 of 139 associated instances:

\medskip
\textbf{Task:} Find the latest log file in a directory matching a given regex pattern; return its path or \texttt{None}. \\
\textbf{Canonical solution:}
\begin{verbatim}
log_files = [f for f in os.listdir(log_dir)
             if re.match(pattern, f)]
log_files = sorted(log_files, key=lambda f:
    os.path.getmtime(os.path.join(log_dir, f)),
    reverse=True)
return os.path.join(log_dir, log_files[0]) \
    if log_files else None
\end{verbatim}

\textbf{Task:} Create a word cloud from text after removing URLs; raise \texttt{ValueError} if no words remain. \\
\textbf{Canonical solution:}
\begin{verbatim}
text = re.sub(r"http[s]?://\S+", "", text)
if not text.strip():
    raise ValueError("No words available...")
wordcloud = WordCloud().generate(text)
plt.imshow(wordcloud)
return wordcloud
\end{verbatim}
\end{rawbox}

\begin{draftbox}[LLM Drafted Skill]
\texttt{name:} re --- Regular Expressions \\
\texttt{description:} Python regular expression library for pattern matching, searching, and text substitution. \\[2pt]
\tcbline

\textbf{Core Functions:}
\begin{itemize}
    \item \texttt{re.match(pattern, string)}: Match pattern at the start of string.
    \item \texttt{re.search(pattern, string)}: Search for pattern anywhere in string.
    \item \texttt{re.findall(pattern, string)}: Return all non-overlapping matches as a list of strings.
    \item \texttt{re.sub(pattern, repl, string)}: Replace all occurrences of pattern.
\end{itemize}

\hfill \textcolor{red}{$\triangleright$ \textit{No behavioral contracts (e.g., \texttt{findall} group semantics, \texttt{match} anchoring)}}

\textbf{Common Patterns:}

Remove URLs: \texttt{re.sub(r"http[s]?://\textbackslash S+", "", text)}

Find all words: \texttt{re.findall(r"\textbackslash w+", text)}

\textbf{Worked Example:}
Remove URLs from text and generate a word cloud:
\texttt{text = re.sub(r"http[s]?://\textbackslash S+", "", text)} \hfill \textcolor{red}{$\triangleright$ \textit{Same task as an input instance}}

[\ldots]
\end{draftbox}

\begin{skillbox}[Golden Skill]
\texttt{name:} re --- Regular Expressions Practical Reference \\
\texttt{description:} Practical guide to re API semantics, findall/sub/match anchoring, case handling, operation ordering, and common text-processing recipes. \\[2pt]
\tcbline

\textbf{Match Anchoring:}
\begin{verbatim}
re.match(r'\w+', text)      # start of string only
re.search(r'\d+', text)     # scans full string
re.fullmatch(r'\w+', text)  # entire string must match
\end{verbatim}

\textbf{\texttt{re.findall} and Capturing Groups:}
\begin{verbatim}
# Without group: returns full matches
re.findall(r'score: \d+', text)
  # -> ['score: 80', 'score: 95']
# With one group: returns group contents only
re.findall(r'score: (\d+)', text)
  # -> ['80', '95']
\end{verbatim}

\textbf{\texttt{re.sub} --- Operation Ordering:}

When steps interact, order matters: remove punctuation \emph{before} introducing underscores, so that \texttt{string.punctuation} (which includes \texttt{\_}) does not remove the new underscores.

\textbf{Word Tokenization Recipes:}
\begin{verbatim}
# Alphabetic words only (drops digits):
re.findall(r'[A-Za-z]+', text)
# Word-character tokens (keeps digits):
re.findall(r'\b\w+\b', text)
\end{verbatim}

[\ldots]
\end{skillbox}

% \noindent\textbf{Revision summary.}
% (1)~The LLM draft lists \texttt{re} functions with textbook-level descriptions but does not specify the behavioral contracts critical for generating correct code: \texttt{re.findall} returns group contents when a capturing group is present (not full matches), \texttt{re.match} is anchored to the string start while \texttt{re.search} scans the full string, and chained \texttt{re.sub} calls interact when later substitutions depend on earlier ones.
% The expert restructured the skill around concrete API contracts with comparison examples and added task-oriented recipes (word tokenization, URL removal, file matching) organized by what they accomplish.
% (2)~The draft's worked example reuses a URL removal task from an input instance.
% The expert replaced it with independently constructed code examples that demonstrate each API contract.

% -------------------------------------------------------------------
\paragraph{Example 2: \texttt{matplotlib} (Data Visualization).}
The \texttt{matplotlib} library appears in 309 instances---the most frequently annotated library---for data visualization tasks; the key challenge lies in returning the correct object type, as different plotting functions produce different return types (\texttt{Axes}, tuples, or ndarrays) that unit tests verify.

\begin{rawbox}[Source Material]
\textbf{Library:} \texttt{matplotlib} --- Data Visualization \\
2 of 309 associated instances:

\medskip
\textbf{Task:} Create a bar chart from a list of category-value pairs; return a \texttt{(DataFrame, Axes)} tuple. \\
\textbf{Canonical solution:}
\begin{verbatim}
df = pd.DataFrame(list_of_pairs,
    columns=["Category", "Value"])
plt.figure(figsize=(10, 5))
sns.barplot(x="Category", y="Value", data=df)
plt.title("Category vs Value")
ax = plt.gca()
return df, ax
\end{verbatim}

\textbf{Task:} Standardize a data matrix, compute row means, and visualize their distribution as a histogram; return a \texttt{(DataFrame, Axes)} tuple. \\
\textbf{Canonical solution:}
\begin{verbatim}
scaler = StandardScaler()
standardized_data = scaler.fit_transform(data_matrix)
df = pd.DataFrame(standardized_data,
    columns=FEATURE_NAMES)
df["Mean"] = df.mean(axis=1)
plt.figure(figsize=(10, 5))
ax = df["Mean"].plot(kind="hist",
    title="Distribution of Means")
return df, ax
\end{verbatim}
\end{rawbox}

\begin{draftbox}[LLM Drafted Skill]
\texttt{name:} matplotlib --- Data Visualization \\
\texttt{description:} Python library for creating plots, charts, and visualizations. \\[2pt]
\tcbline

\textbf{Basic Plotting Functions:}
\begin{itemize}
    \item \texttt{plt.plot(x, y)}: Create a line plot.
    \item \texttt{plt.bar(x, height)}: Create a bar chart.
    \item \texttt{plt.hist(data, bins)}: Create a histogram.
    \item \texttt{plt.scatter(x, y)}: Create a scatter plot.
\end{itemize}

\textbf{Customization:}
\texttt{plt.title("Title")}, \texttt{plt.xlabel("X")}, \texttt{plt.ylabel("Y")}, \texttt{plt.figure(figsize=(w, h))}.

\hfill \textcolor{red}{$\triangleright$ \textit{No return type specifications---the most common source of test failures}}

\textbf{Worked Example:}
Create a bar chart from a DataFrame:
\begin{verbatim}
plt.figure(figsize=(10, 5))
sns.barplot(x="Category", y="Value", data=df)
ax = plt.gca()
return df, ax
\end{verbatim}
\hfill \textcolor{red}{$\triangleright$ \textit{Same task as an input instance}}

[\ldots]
\end{draftbox}

\begin{skillbox}[Golden Skill]
\texttt{name:} matplotlib --- Plotting API, Return Types, and Composition Recipes \\
\texttt{description:} Guide for correct matplotlib API usage: return type contracts, OOP vs state-machine style, histogram rwidth, bar chart positioning, and common pitfalls. \\[2pt]
\tcbline

\textbf{Return Type Quick Reference:}

\medskip
{\small
\begin{tabular}{ll}
\toprule
\textbf{Expression} & \textbf{Actual return type} \\
\midrule
\texttt{fig, ax = plt.subplots()} $\to$ \texttt{ax} & Axes \\
\texttt{plt.bar(labels, vals)} & BarContainer \\
\texttt{plt.hist(data)} & (n, bins, patches) tuple \\
\texttt{df.plot()} / \texttt{series.plot()} & Axes \\
\texttt{df.boxplot()} & Axes \\
\texttt{ax.pie(\ldots)} & (wedges, texts) tuple \\
\bottomrule
\end{tabular}
}

\medskip
\textbf{OOP Style} (use when returning Axes):
\begin{verbatim}
fig, ax = plt.subplots()
ax.plot(x, y, label="series")
ax.set_title("Title")
return ax
\end{verbatim}

\textbf{Bar Charts with Numeric Positions:}
\begin{verbatim}
indexes = np.arange(len(labels))
ax.bar(indexes, values)
ax.set_xticks(indexes)       # positions first
ax.set_xticklabels(labels)   # then labels
\end{verbatim}
Omitting \texttt{set\_xticks} before \texttt{set\_xticklabels} raises a \texttt{ValueError}.

[\ldots]
\end{skillbox}

\noindent\textbf{Revision summary.}
(1)~The LLM draft describes matplotlib's plotting functions at a conceptual level but omits return type specifications---the single most common source of test failures in BigCodeBench.
Functions like \texttt{plt.hist()} return a \texttt{(n, bins, patches)} tuple rather than an Axes object, \texttt{df.boxplot()} does not return Axes, and \texttt{ax.pie()} returns a tuple of wedges and texts.
Models that follow the draft's function-level descriptions produce code that returns incorrect types and fails unit tests.
The expert added a return type reference table and restructured the skill around the OOP vs.\ state-machine distinction, with correct return patterns for each plotting idiom.
(2)~The draft's worked example reuses a bar chart task from an input instance.
The expert replaced it with newly constructed code examples that exercise different return type patterns.

\section{Implementation Details}
\label{sec:appendix-inference}

This appendix provides the prompt templates used to implement the skill-use strategies introduced in~\S\ref{sec:experimental-setup}. 
These templates specify how retrieved skills are exposed to the model during inference. 
Unless otherwise stated, \texttt{\{original user prompt\}} denotes the original task prompt from the corresponding benchmark.

\subsection{Full-Skill Injection}
\label{sec:appendix-prepend}

When one or more skills are provided, either by retrieval or by Oracle annotation, their full content is prepended to the user prompt:

\begin{promptbox}[Full-Skill Injection]
\ttfamily
Relevant Skill:\\
\{skill content\}\\[2pt]
\{original user prompt\}
\end{promptbox}

\noindent When multiple skills are available, e.g., in Oracle mode for multi-label datasets such as CHAMP~\cite{mao2024champ} and BigCodeBench~\cite{zhuo2024bigcodebench}, they are concatenated with a horizontal rule delimiter (\texttt{-{}-{}-}).

\subsection{LLM Selection}
\label{sec:appendix-select}

For LLM Selection, the model is first given the task query alongside a numbered list of candidate skills, where each candidate is represented by its name and description. 
The model is asked to output only the number of the most relevant skill:

\begin{promptbox}[LLM Selection prompt]
\ttfamily
Given the following problem, select the ONE most relevant skill. Output ONLY the skill number.\\[2pt]
Problem:\\
\{query\}\\[2pt]
Skills:\\{}
[1] Bayes' Theorem: Posterior probability computation using prior and likelihood\\{}
[2] Law of Total Probability: Marginal probability via exhaustive partitioning\\
...\\[2pt]
Most relevant skill number:
\end{promptbox}

\noindent The selected skill's full content is then prepended to the task prompt using the same format as Full-Skill Injection in~\S\ref{sec:appendix-prepend}.

\subsection{Progressive Disclosure}
\label{sec:appendix-agentic}

For Progressive Disclosure, the system prompt is augmented with a compact catalog of candidate skills and an instruction that allows the model to load full skill content on demand:

\begin{promptbox}[Progressive Disclosure prompt (appended to system prompt)]
\ttfamily
You have access to a skill library. Each skill provides precise methodology and step-by-step procedures for a specific problem type --- these often contain critical details that general knowledge may miss.\\[2pt]
To use a skill, write on its own line:\\
LOAD\_SKILL: <index>\\[2pt]
For example: LOAD\_SKILL: 0\\[2pt]
You will receive the skill's full content and can then apply the methodology to solve the problem.\\[2pt]
Available skills:\\
0 --- Bayes' Theorem --- Posterior probability computation using prior and likelihood\\
1 --- Law of Total Probability --- Marginal probability via exhaustive partitioning\\
...
\end{promptbox}

\noindent When the model issues a \texttt{LOAD\_SKILL} action, the framework injects the full content of the corresponding skill and prompts the model to continue. 
The model may load multiple skills across successive turns, up to 10 rounds.

\end{document}